\documentclass[sigconf]{acmart}

\usepackage{booktabs} 

\usepackage{amsfonts,dsfont}
\usepackage{amssymb}
\usepackage{amsmath}
\usepackage{graphicx}
\usepackage[]{natbib}
\usepackage[caption=false,font=footnotesize,labelfont=sf,textfont=sf]{subfig}
\usepackage{color}

\usepackage{algorithm}
\usepackage{algorithmic}

\newlength\myindent
\newcommand\bindent{%
	\begingroup
	\setlength{\itemindent}{\myindent}
	\addtolength{\algorithmicindent}{\myindent}
}
\newcommand\eindent{\endgroup}

\DeclareMathOperator*{\argmin}{arg\,min}
\DeclareMathOperator*{\argmax}{arg\,max}

\definecolor{britishracinggreen}{rgb}{0.0, 0.26, 0.15}

\setcopyright{rightsretained}

\acmDOI{}

\acmISBN{}

\acmConference[KDD 2017 Workshop on Interactive Data Exploration and Analytics (IDEA'17)]{}{August 14th, 2017}{Halifax, Nova Scotia, Canada} 
\acmYear{2017}
\copyrightyear{2017}

\acmPrice{}

\begin{document}
\title{Incorporating Feedback into Tree-based Anomaly Detection}

\author{Shubhomoy Das}
\affiliation{%
  \institution{Oregon State University}
  \city{Corvallis} 
  \state{Oregon} 
  \postcode{97330}
}
\email{dassh@oregonstate.edu}

\author{Weng-Keen Wong}
\affiliation{%
  \institution{Oregon State University}
\city{Corvallis} 
\state{Oregon} 
\postcode{97330}
}
\email{wongwe@oregonstate.edu}

\author{Alan Fern}
\affiliation{%
	\institution{Oregon State University}
	\city{Corvallis} 
	\state{Oregon} 
	\postcode{97330}
}
\email{Alan.Fern@oregonstate.edu}

\author{Thomas G. Dietterich}
\affiliation{%
	\institution{Oregon State University}
	\city{Corvallis} 
	\state{Oregon} 
	\postcode{97330}
}
\email{tgd@oregonstate.edu}

\author{Md Amran Siddiqui}
\affiliation{%
	\institution{Oregon State University}
	\city{Corvallis} 
	\state{Oregon} 
	\postcode{97330}
}
\email{siddiqmd@oregonstate.edu}

\renewcommand{\shortauthors}{Das et al.}

\begin{abstract}
Anomaly detectors are often used to produce a ranked list of statistical anomalies, which are examined by human analysts in order to extract the actual anomalies of interest. Unfortunately, in real-world applications, this process can be exceedingly difficult for the analyst since a large fraction of high-ranking anomalies are false positives and not interesting from the application perspective. 
In this paper, we aim to make the analyst's job easier by allowing for analyst feedback during the investigation process. Ideally, the feedback influences the ranking of the anomaly detector in a way that reduces the number of false positives that must be examined before discovering the anomalies of interest. 
In particular, we introduce a novel technique for incorporating simple binary feedback into tree-based anomaly detectors. We focus on the Isolation Forest algorithm as a representative tree-based anomaly detector, and show that we can significantly improve its performance by incorporating feedback, when compared with the baseline algorithm that does not incorporate feedback. Our technique is simple and scales well as the size of the data increases, which makes it suitable for interactive discovery of anomalies in large datasets.
\end{abstract}

%
%



\maketitle

\section{Introduction}
We define an \textit{anomaly} as a data instance generated by a different process than the process generating the nominal data. On the other hand, we define an \textit{outlier} as a data instance that has low likelihood according to a model. Anomaly detectors are in general very good at detecting outliers. However, not all outliers are anomalies. Some outliers are statistical noise, while others might not interest the end-users. A class of the state-of-the-art anomaly detectors dependent on unsupervised tree-based methods \cite{liu:08, tan:2011, chen:2015, wu2014rs} are not naturally immune to this problem. These detectors usually partition the feature space into multiple (sometimes overlapping and hierarchical) regions and assign scores to each region individually. When the scores computed for some of the regions do not reflect their true relevance to the user's notion of an anomaly, it creates a semantic mismatch between what the user considers an anomaly and what the algorithm considers an outlier. In order to avoid this mismatch, we need expert-feedback to make outliers more in line with expert's idea of an anomaly.

Active Anomaly Discovery (AAD) \cite{Das:16} is one of the most recent methods for incorporating analyst-feedback into an ensemble of anomaly detectors. In this paper, we show that tree-based anomaly detectors can also be treated as \textit{ensembles} such that we can incorporate feedback into them by employing AAD. We present an implementation of this concept in the specific context of the tree-based anomaly detector Isolation Forest \cite{liu:08}, which is competitive with other state-of-the-art anomaly detectors \cite{liu:08, emmott:2015}. One advantage of the proposed approach is that it allows incorporating feedback at a finer level than simply combining the outputs of multiple detectors linearly.

In Section~\ref{sec:treebased}, we present our view of the general structure of tree-based anomaly detectors, and illustrate this view with Isolation Forest as an example. Section~\ref{sec:reweight} presents an overview of AAD and then extends AAD to incorporate feedback into the Isolation Forest. We refer to this new algorithm as \textit{IF-AAD}. Section~\ref{sec:experiments} presents quantitative empirical results on eight benchmark datasets and provides a visualization of the feedback process in order to gain further insight into how the feedback affects which instances are queried. Finally, we summarize the contributions and results in Section~\ref{sec:conclusion}.

\section{Tree-based Anomaly Detectors}
\label{sec:treebased}
\begin{table*}[!t]
	\caption{Node weights for tree-based algorithms. \label{tab:tree_based}}
	{\begin{tabular}{l c l}
			{\bf Name} & {\bf Internal node weight} & {\bf Leaf node weight} \\ \hline
			Isolation Forest \cite{liu:08} & $-1$ & $-1$
			\\ \hline
			HS-Trees \cite{tan:2011} & $0$ & anomaly score as defined in \citet{tan:2011}
			\\ \hline
			RS-Forest \cite{wu2014rs} & $0$ & anomaly score as defined in \citet{wu2014rs}
			\\ \hline
			RPAD (`AVG' variant) \cite{siddiqui:2016} & normalized pattern frequency \cite{siddiqui:2016} & normalized pattern frequency \cite{siddiqui:2016}
			\\ \hline
			Random Projection Forest \cite{chen:2015} & log-probability at the node \cite{chen:2015} & log-probability at the leaf \cite{chen:2015}
			\\ \hline
	\end{tabular}}
\end{table*}


We consider an anomaly detection setting where an anomaly detector is used to assign anomaly scores to data instances, which are assumed to be feature vectors in $\mathcal{R}^n$. The instances can then be presented to an analyst in ranked order, starting with the most anomalous instance.   
Our work is motivated by the observation that a number of state-of-the-art anomaly detectors are based on decision-tree ensembles, or forests. The internal nodes of each tree correspond to threshold tests on selected features. Thus, a given instance $x$ will follow a unique path from the root to a leaf in each tree. 

Each tree node $\nu$ in the tree-based anomaly detector stores a real-valued weight $w_{\nu}$, which is used to calculate the anomaly scores. The anomaly score of an instance $x$ is simply equal to the average over weights of all tree nodes that the instance follows in the forest. Note that each node in the forest can be viewed as defining a distinct volume in $\mathcal{R}^n$ and thus the total score is a combination of the weights of these overlapping volumes. 

Despite the simplicity of this anomaly detection structure, a number of state-of-the-art algorithms can be represented as a particular choice of weight values and methods to construct the trees (generally highly randomized trees). Table \ref{tab:tree_based} illustrates the weight values that correspond to a number of algorithms. As one example and as described in detail below, the Isolation Forest \cite{liu:08} algorithm assigns a constant weight of $-1$ to all tree nodes.\footnote{This assumes the trees are grown to a depth where instances are isolated. Otherwise the leaf nodes would have alternative weights that depend on the amount of data arriving at each leaf.} The anomaly score then evaluates to be the average path length traversed by an instance across trees in the forest. 

As another example, the HS-Trees\cite{tan:2011} algorithm, assigns a weight of $\nu_r \times 2^{\nu_k}$ to each \textbf{leaf} node $\nu$, where $\nu_r$ is the number of training instances at the node $\nu$, and $\nu_k$ is the node's depth. In addition, HS-Trees assigns weight $0$ to all non-leaf nodes. Thus, in this case the anomaly score of an instance $x$ is the average of the weights at leaves it reaches. 

In practice, there is no uniformly best anomaly detector (or equivalently, a fixed setting of the weights) across the possible applications. Rather, the best performing detector for a given application will depend on how well a detector's notion of ``outlier'' matches the analyst's notion of ``interesting anomaly''. This is difficult to predict for a given application. Further, it is unlikely that any of the weight settings corresponding to state-of-the-art detectors will be optimal for a given application when considering the entire range of possible weight settings. 


The above motivates incorporating user feedback during use of an anomaly detector to attempt to tune the weights toward the ideal application-specific detector. In Section~\ref{sec:experiments}, we show that this approach often increases the number true anomalies discovered within a particular budget of instances that can be examined by an analyst. In this paper, we treat Isolation Forest as a representative tree-based anomaly detector, and explain our method for incorporating feedback, where the detector is initialized to the Isolation Forest weights. Below we describe in detail the Isolation Forest algorithm for concreteness and illustrate how it is easily captured in our tree-based anomaly detection framework. 



\subsection{Isolation Forest (IF)}
\label{sec:if}
\begin{figure*}[!t]
	\centering
	\subfloat[Isolation Tree Construction]{\includegraphics[width=4.0in]{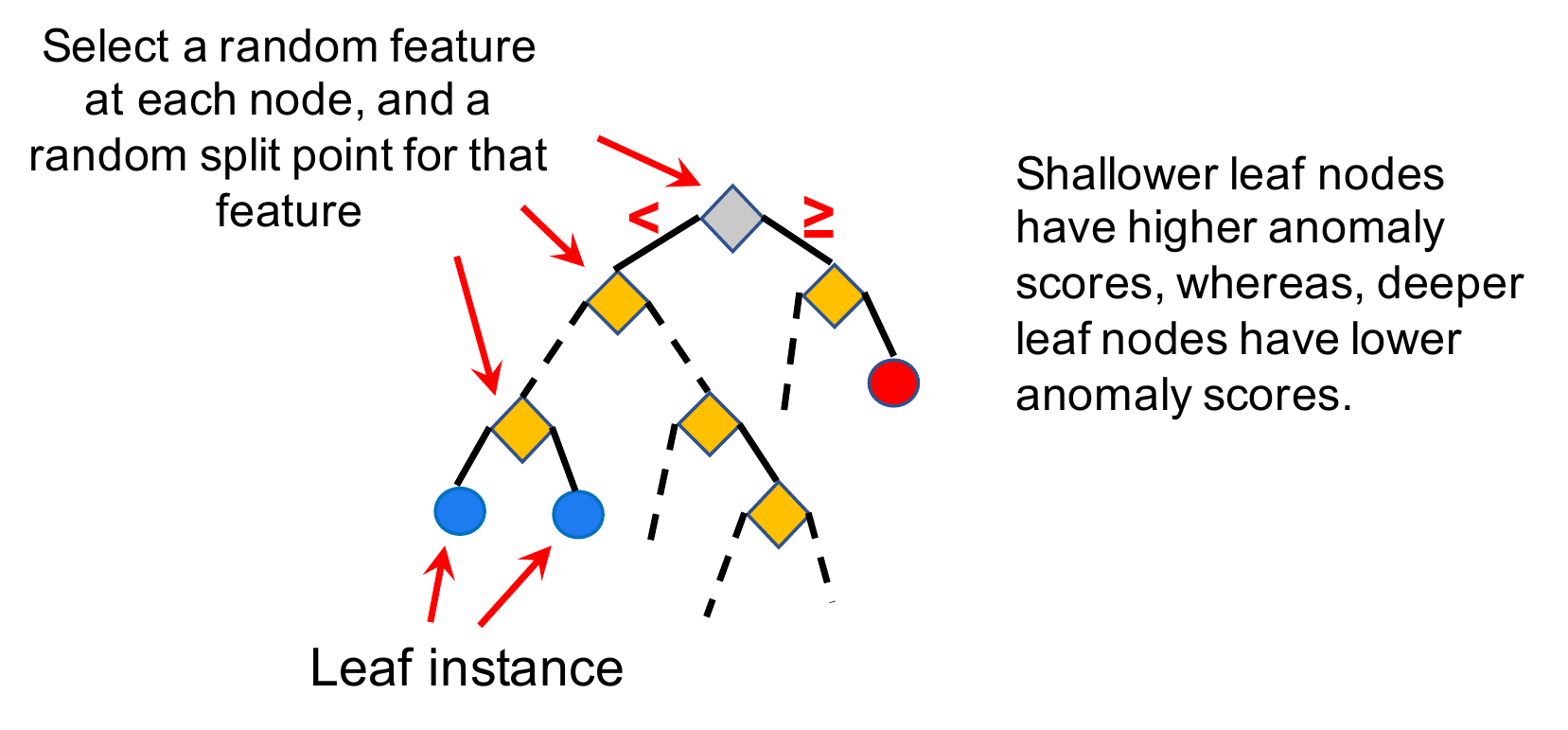}%
		\label{fig:ifor_construction}} \\
	\subfloat[Synthetic dataset]{\includegraphics[width=2.0in]{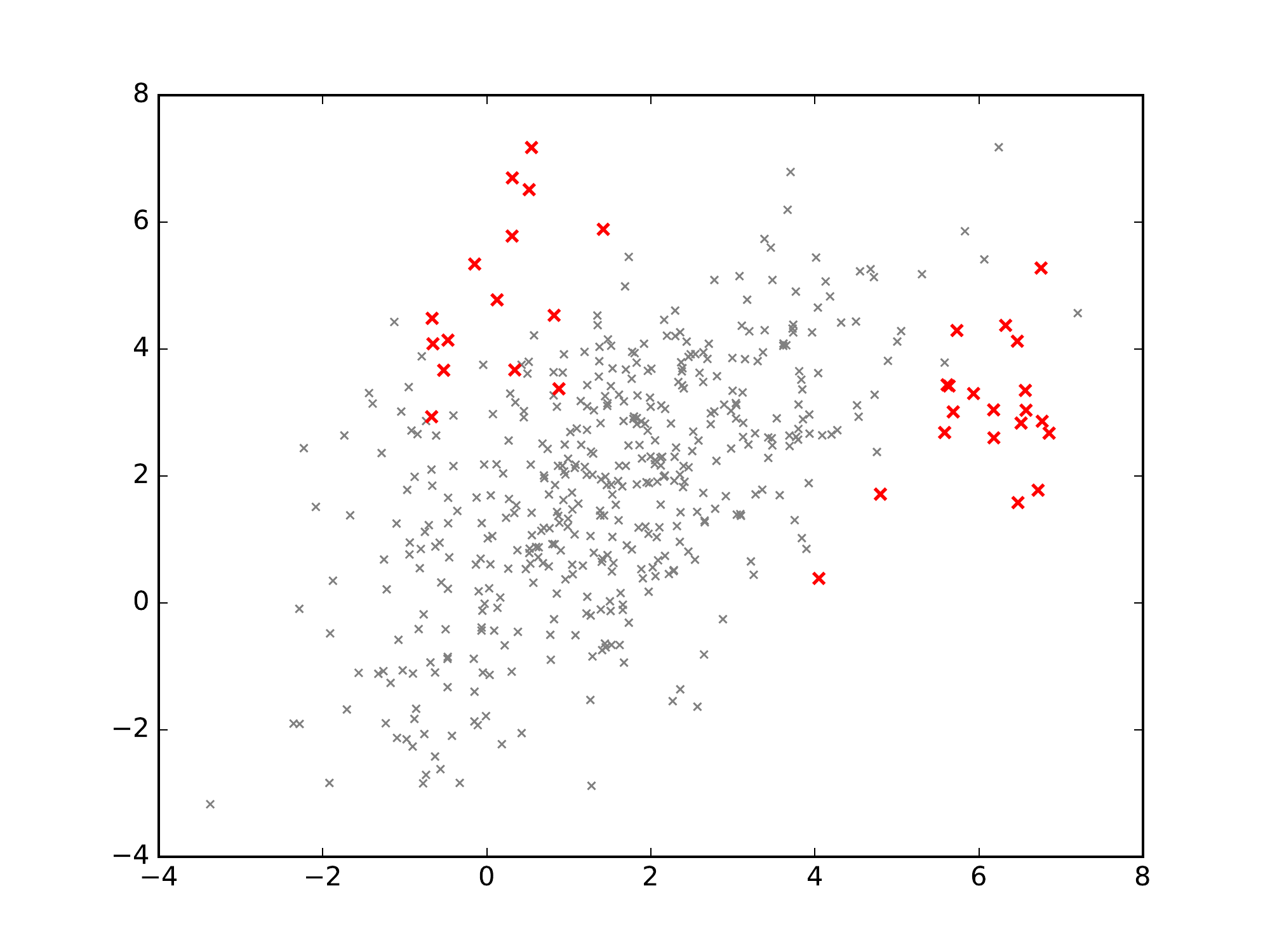}%
		\label{fig:synthetic_data}}
	\subfloat[1 Tree]{\includegraphics[width=2.0in]{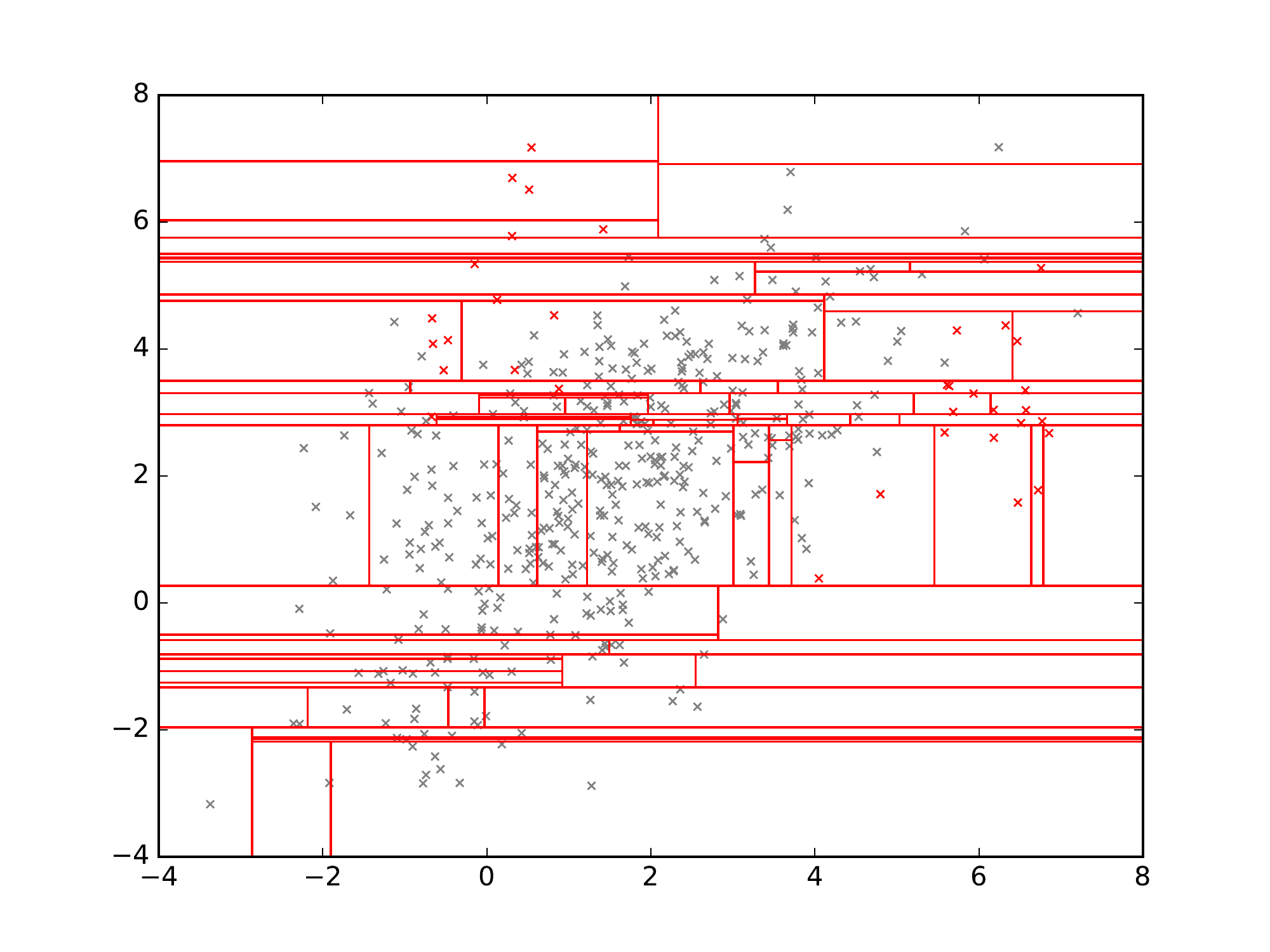}%
		\label{fig:ifor_1_trees}}
	\subfloat[Contours with 1 Tree]{\includegraphics[width=2.0in]{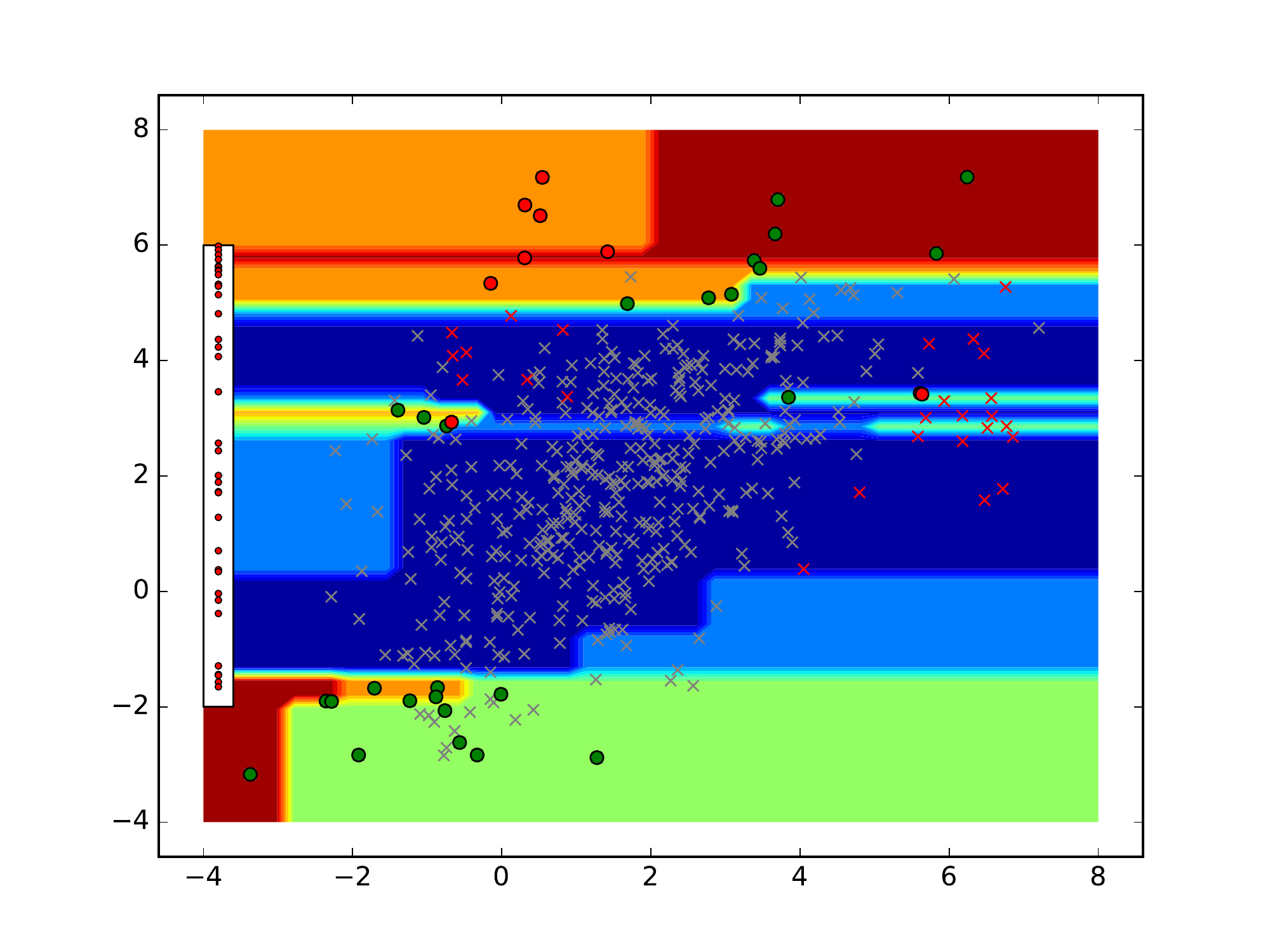}%
		\label{fig:ifor_contours_1_trees}}
	\caption{Random trees in Isolation Forest (IF) for synthetic data. The points in \textcolor{red}{red} are true anomalies; points in gray are true nominals. Figure~\ref{fig:ifor_1_trees} shows the leaf node regions for a single tree generated by random IF splits. Figure~\ref{fig:ifor_contours_1_trees} shows the contours of anomaly scores assigned to the nodes of this tree. Deeper \textcolor{red}{red} means more anomalous; deeper \textcolor{blue}{blue} means more nominal. The \textcolor{red}{red} circles are the true anomalies among the top ranked 35 instances. The \textcolor{britishracinggreen}{green} circles are the true nominals among the top ranked 35 instances. The left sidebar in Figure~\ref{fig:ifor_contours_1_trees} shows the ranking of true anomalies (\textcolor{red}{red} dots). Ideally, true anomalies should be near the top on this bar.}
	\label{fig:ifor_contours_trees}
\end{figure*}

Isolation Forest (IF) \cite{liu:08} comprises of a set of $t$ trees denoted by $\mathbf{T} = \{T_1, ..., T_t\}$ constructed in a randomized manner as outlined in Algorithm~\ref{alg:iforest}, and illustrated in Figure~\ref{fig:ifor_construction}. Each tree is constructed from the root to leaves by randomly partitioning the data at each node by selecting a feature and a threshold both at uniformly random. The trees are grown until each instance is isolated in a leaf. IF is based on the idea that anomalous instances are few, and they are well-separated from clusters of nominal instances in the feature space. Because of this, anomalous instances very quickly reach leaf nodes through random partitioning. On the other hand, nominal instances, which form dense clusters, require many more splits to finally reach leaf nodes. Therefore, the length of the path traversed by an instance from the root node to the leaf, also known as the isolation depth, is shorter (on average) for anomalous instances than it is for nominal instances. The anomaly score assigned to an instance is simply the average isolation depth across the forest. 


It is straightforward now to describe IF as a particular way of setting the weights of a tree-based anomaly detector. In particular, the weight of each node $\nu$ is $w_{\nu} = -1$ (constant). Given an instance $x$, it is easy to see that the anomaly score assigned by the tree-based detector is simply negative of the average number of nodes on paths traversed by $x$ in the forest, i.e. negative of the average isolation depth. Note that, the main purpose to make scores negative is to ensure that higher scores indicate more anomalous and lower scores indicate more nominal.



In order to describe our algorithm for feedback, it is convenient to view the score assigned by the detector as a linear score function. To do this, for each tree node $\nu$ define an indicator feature $z_{\nu} \in \{0, 1\}$. The anomaly score is then simply the dot product of feature and weight vectors, that is, $$\mbox{score}(x) = z\cdot w,$$
where the dimension of each vector is the number of nodes in the forest. 


Figure~\ref{fig:ifor_contours_trees} illustrates the anomaly score contours for IF with a single tree on synthetic data.
The anomaly score contours in Figure~\ref{fig:ifor_contours_1_trees} show that a single isolation tree is not very informative. However, if we increase the number of trees in the ensemble, their combined scores can be fairly accurate even without feedback. This is illustrated in Figure~\ref{fig:ifor_iter_00} where the number of trees is $100$.

\begin{algorithm}[h!]
	\caption{Generating randomized trees in Isolation Forest}
	\label{alg:iforest}
	\begin{algorithmic}
		\STATE \textbf{Input:} $\mathbf{D}$, sub-sample size: $N$, number of trees: $t$
		\STATE $\mathbf{T} = \emptyset$
		\FOR{$i = 1 ... t$}
		\STATE Let $S_i$ = a sub-sample of $N$ instances from $\mathbf{D}$
		\STATE Build tree $T_i$ as follows, by starting with all instances in $S_i$ at the root node:
		\bindent
		\STATE Let $U \subseteq S_i$ be the set of instances at the current node
		\IF{$|U| == 1$}
		\RETURN
		\ELSE
		\STATE Let $f$ be a feature sampled at random
		\STATE Let $f_{min}$ = min. value of $f$ across all instances in $U$
		\STATE Let $f_{max}$ = max. value of $f$ across all instances in $U$
		\STATE Let $p_f$ = value sampled unif. random in $[f_{min}, f_{max}]$
		\STATE Partition $U$ into two parts on the basis of $p_f$ and recurse on both partitions
		\ENDIF
		\eindent
		\STATE $\mathbf{T} = \mathbf{T} \cup T_i$
		\ENDFOR
	\end{algorithmic}
\end{algorithm}

\section{Re-weighting Tree Partitions}
\label{sec:reweight}

We now describe our approach for adjusting the weights in the above score function based on feedback from the analyst. 

\subsection{Active Anomaly Discovery (AAD)}
\label{sec:aad}
AAD is an algorithm (Algorithm~\ref{alg:fbloop}) that tries to maximize the number of true anomalies presented to the analyst in an interactive feedback loop. It assigns an anomaly score to each instance such that a higher score means more anomalous. The instances are internally ranked in descending order of the scores. In each feedback iteration, AAD presents the most anomalous instance to the analyst and asks for its true label, either \emph{anomalous} or \emph{nominal}. In prior work, the AAD algorithm was developed to learn the weighting among an ensemble of anomaly detectors, in particular ensembles produced by the LODA \cite{pevny:2015} anomaly detector. Here we show that the same approach can be used to re-weight nodes within the trees of a forest.  

Assume that we have a dataset instance ${\mathbf H} = \{{\mathbf z}_1, ..., {\mathbf z}_n\}$, where ${\mathbf z}_i \in \mathds{R}^M$. Note that here we think of the instances as being represented by the vector of indicator features corresponding to tree nodes. When the label is known for an instance ${\mathbf z}_i$, we will denote the label by $y_i \in \{anomaly, nominal\}$. Let ${\mathbf H}_F \subseteq {\mathbf H}$ be the set of instances for which the analyst has already provided feedback, $\mathbf{H}_A \subseteq {\mathbf H}_F$ be the set of labeled anomalies, and let $\mathbf{H}_N \subseteq {\mathbf H}_F$ be the set of labeled nominals. The anomaly score of an instance ${\mathbf z}$ is $\text{score}({\mathbf z}) = {\mathbf z} \cdot {\mathbf w}$, and our goal is to learn the weights $w$ that will most likely rank the true anomalies near the top. 

The AAD algorithm takes a quantile parameter as input $\tau \in [0, 1]$. The instance that has the $\tau$-th ranked score (in descending order) is denoted by ${\mathbf z}_{\tau}$, and its corresponding score is denoted by $q_{\tau}$.
The weight vector ${\mathbf w}$ must ensure that scores of labeled anomalies ${\mathbf z} \in {\mathbf H}_A$ are higher than $q_{\tau}$ while, at the same time, the scores of labeled nominals ${\mathbf z} \in {\mathbf H}_N$ are lower than $q_{\tau}$. Additionally, AAD adds soft pairwise constraints which encourage every labeled anomaly to have a higher score than every labeled nominal under the new weights that are learned.

The weight vector ${\mathbf w}$ is learned through a constrained optimization problem (described below). This problem is the same as the one introduced for the original AAD algorithm \cite{Das:16}, except for the following differences: 
\begin{enumerate}
	\item Instead of introducing all pairwise constraints between anomalies and nominals, we only add constraints relative to the current $\tau$-th ranked instance. We found that this change does not degrade the accuracy of AAD in detecting anomalies, but makes the computation significantly faster.
	\item Since the pairwise constraints are `soft', each violated constraint is 
	multiplied by a slack penalty term $C_{\xi}$. We can then re-formulate the objective by adding additional terms to the loss function that correspond to the constraints. This allows optimization by gradient descent, which is helpful when the number of features is very high --- as will be the case in our proposed algorithm.
\end{enumerate}

Before formulating the optimization problem, we first define the following hinge loss $\ell(q, {\mathbf w}; ({\mathbf z}_i, y_i))$:
\begin{align}
\ell(q, {\mathbf w}; ({\mathbf z}_i, y_i)) = \qquad \qquad \qquad \qquad \qquad \qquad \qquad \qquad & \nonumber \\
\left \{ 
\begin{array}{lr}
0 & {\mathbf w}\cdot {\mathbf z}_i \ge q \text{ and $y_i=$\textit{`anomaly'}} \\
0 & {\mathbf w}\cdot {\mathbf z}_i < q \text{ and $y_i=$\textit{`nominal'} } \\
(q - {\mathbf w}\cdot {\mathbf z}_i) & {\mathbf w}\cdot {\mathbf z}_i < q \text{ and  $y_i=$\textit{`anomaly'}} \\
({\mathbf w}\cdot {\mathbf z}_i - q) & {\mathbf w}\cdot {\mathbf z}_i \ge q \text{ and $y_i=$\textit{`nominal'}}
\end{array} 
\right. & \nonumber \\
& \label{eqn:aatploss}
\end{align}

The modified unconstrained optimization problem for learning the optimal weights is then formulated as:
\begin{align}
{\mathbf w}^{(t)} =& \argmin_{{\mathbf w}, {\mathbf \xi}} \frac{C_A}{|{\mathbf H}_A|}\left ( \sum_{{\mathbf z}_i \in {\mathbf H}_A}\ell(\hat{q}_{\tau}( {{\mathbf w}^{(t-1)}}), {\mathbf w}; ({\mathbf z}_i, y_i)) \right ) \nonumber \\
& \qquad \qquad + \frac{1}{|{\mathbf H}_N|}\left ( \sum_{{\mathbf z}_i \in {\mathbf H}_N}\ell(\hat{q}_{\tau}({{\mathbf w}^{(t-1)}}), {\mathbf w}; ({\mathbf z}_i, y_i)) \right ) \nonumber \\ 
& \qquad \qquad + \frac{C_{\xi}}{|{\mathbf H}_A|}\left ( \sum_{{\mathbf z}_i \in {\mathbf H}_A}\ell({\mathbf z}_{\tau}^{(t-1)} \cdot {\mathbf w}, {\mathbf w}; ({\mathbf z}_i, y_i)) \right ) \nonumber \\ 
& \qquad \qquad + \frac{C_{\xi}}{|{\mathbf H}_N|}\left ( \sum_{{\mathbf z}_i \in {\mathbf H}_N}\ell({\mathbf z}_{\tau}^{(t-1)} \cdot {\mathbf w}, {\mathbf w}; ({\mathbf z}_i, y_i)) \right ) \nonumber \\ 
& \qquad \qquad + \|{\mathbf w} - {\mathbf w}_p\|^2 \label{eqn:preflearn_aatp}
\end{align}
where, $\mathbf{w}_p = \frac{\mathbf{w}_U}{\|\mathbf{w}_U\|}=[\frac{1}{\sqrt{m}},\ldots,\frac{1}{\sqrt{m}}]^T$, ${\mathbf z}_{\tau}^{(t-1)}$ and $\hat{q}_{\tau}({{\mathbf w}^{(t-1)}})$ are computed by ranking anomaly scores with ${\mathbf w} = {\mathbf w}^{(t-1)}$. $C_A$ and $C_{\xi}$ are constant weight hyper-parameters. When $C_A$ is set to a value larger than $1$, as is typically the case, it causes the hinge loss for anomalies in $\mathbf{H}_A$ to be higher than those associated with nominals. $C_{\xi}$ encourages a) the scores of anomalies in $\mathbf{H}_A$ to be higher than that of the $\tau$-th ranked instance from the previous iteration, and b) the scores of nominals in $\mathbf{H}_N$ to be lower than that of the $\tau$-th ranked instance from the previous iteration.

We apply gradient descent to learn the optimal weights ${\mathbf w}$ for Equation~\ref{eqn:preflearn_aatp}, in Line~\ref{alg:fbloop:weightupdate} of Algorithm~\ref{alg:fbloop}.

\begin{algorithm}
	\caption{Active Anomaly Discovery (AAD)}
	\label{alg:fbloop}
	\begin{algorithmic}[15]
		\STATE \textbf{Input:} Dataset ${\mathbf H}$, budget $B$
		\STATE Initialize the weights ${\mathbf w}^{(0)} = \{\frac{1}{\sqrt{m}}, ..., \frac{1}{\sqrt{m}}\}$
		\STATE Set $t=0$
		\STATE Set ${\mathbf H}_A = {\mathbf H}_N = \emptyset$
		\WHILE{$t \leq B$}
		\STATE $t = t + 1$
		\STATE Set ${\mathbf a} = {\mathbf H} \cdot {\mathbf w}$ (i.e., ${\mathbf a}$ is the vector of anomaly scores)
		\STATE Let ${\mathbf z}_i$ = instance with highest anomaly score (where $i = \argmax_{i}(a_i)$)
		\STATE Get feedback $\{\text{\textit{`anomaly'}} / \text{\textit{`nominal'}}\}$ on ${\mathbf z}_i$
		\IF{${\mathbf z}_i$ is \textit{anomaly}}
		\STATE ${\mathbf H}_A = \{{\bf z}_i\} \cup {\mathbf H}_A$
		\ELSE
		\STATE ${\mathbf H}_N = \{{\bf z}_i\} \cup {\mathbf H}_N$
		\ENDIF
		\STATE ${\mathbf w}^{(t)}$ = compute new weights; normalize $\|{\mathbf w}^{(t)}\|=1$ \label{alg:fbloop:weightupdate}
		\ENDWHILE
	\end{algorithmic}
\end{algorithm}

\subsection{Re-weighting IF Partitions (IF-AAD)}
\label{sec:if_aad}
\begin{figure*}[!t]
	\centering
	\subfloat[Initial]{\includegraphics[width=2.0in]{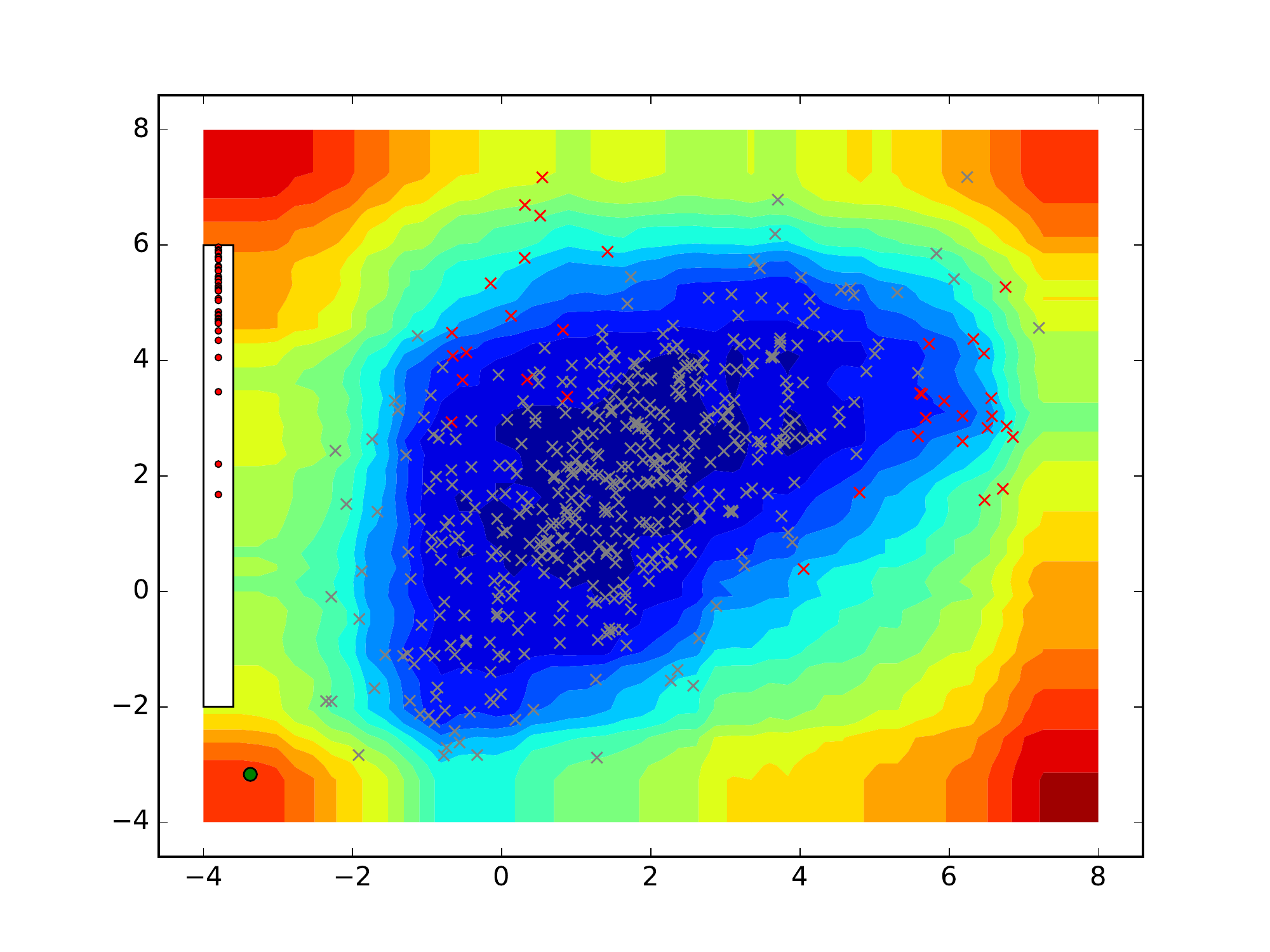}%
		\label{fig:ifor_iter_00}}
	\subfloat[8 Iterations]{\includegraphics[width=2.0in]{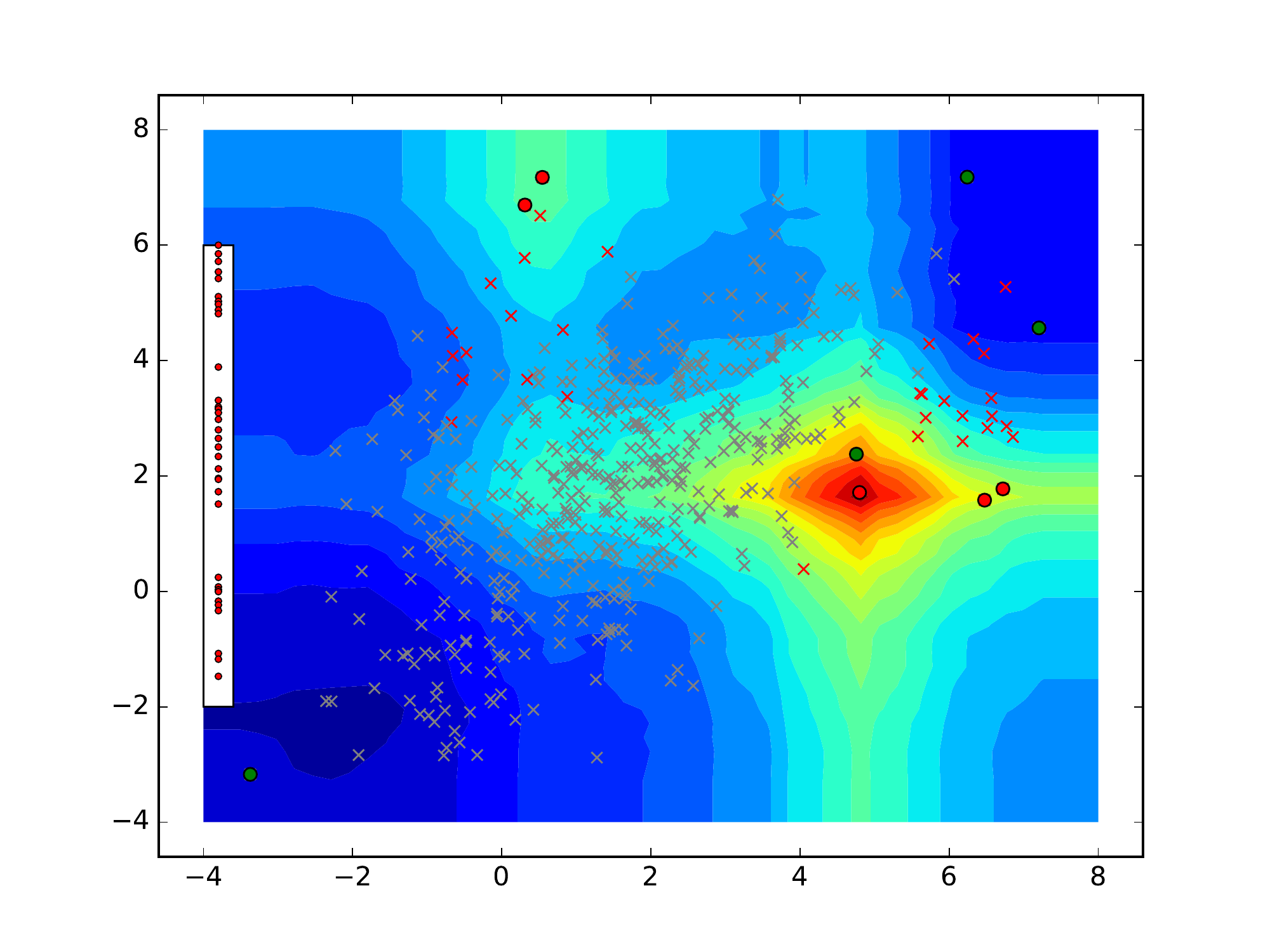}%
		\label{fig:ifor_iter_08}}
	\subfloat[16 Iterations]{\includegraphics[width=2.0in]{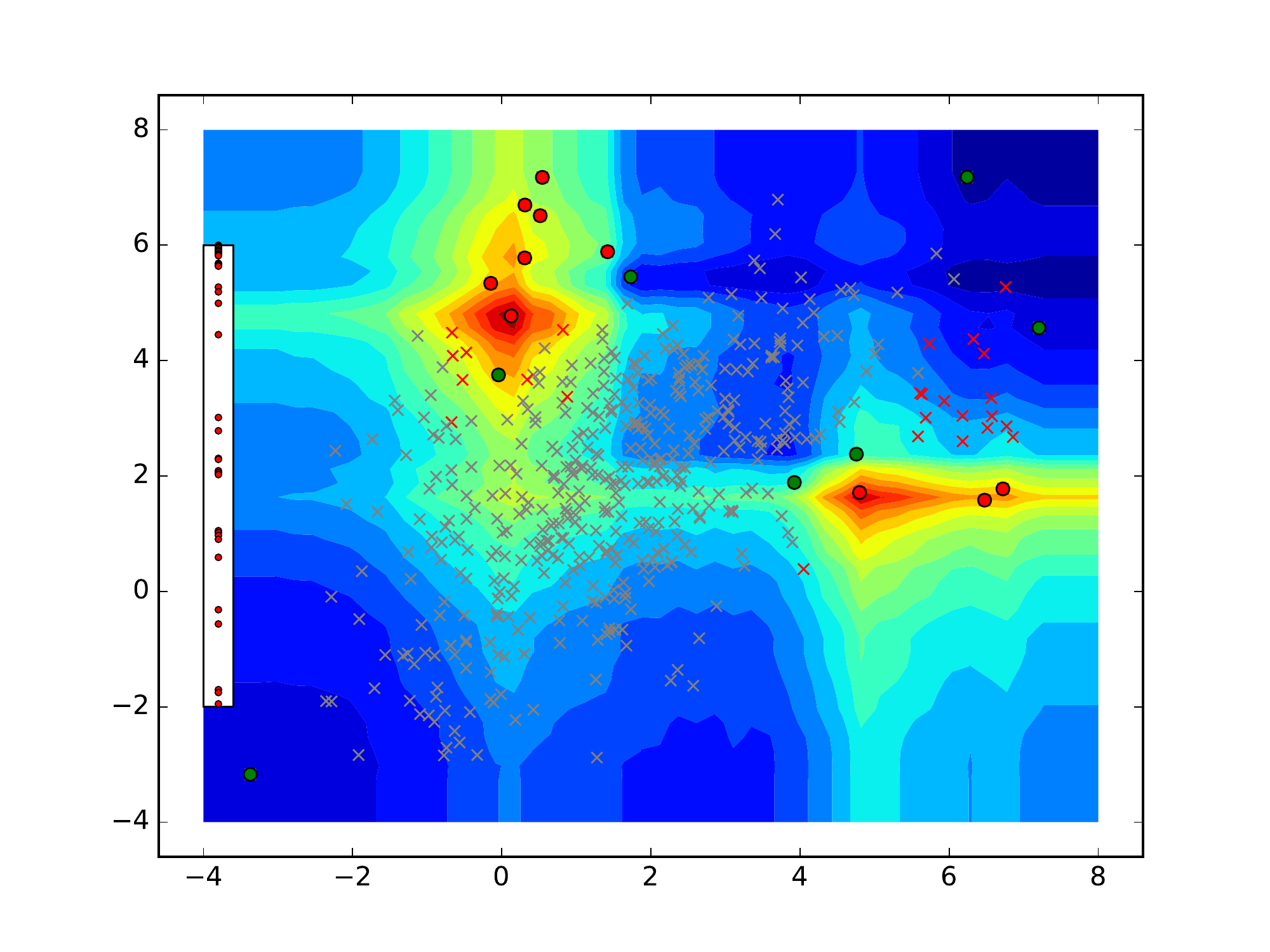}%
		\label{fig:ifor_iter_16}} \\[-2ex]
	\subfloat[24 Iterations]{\includegraphics[width=2.0in]{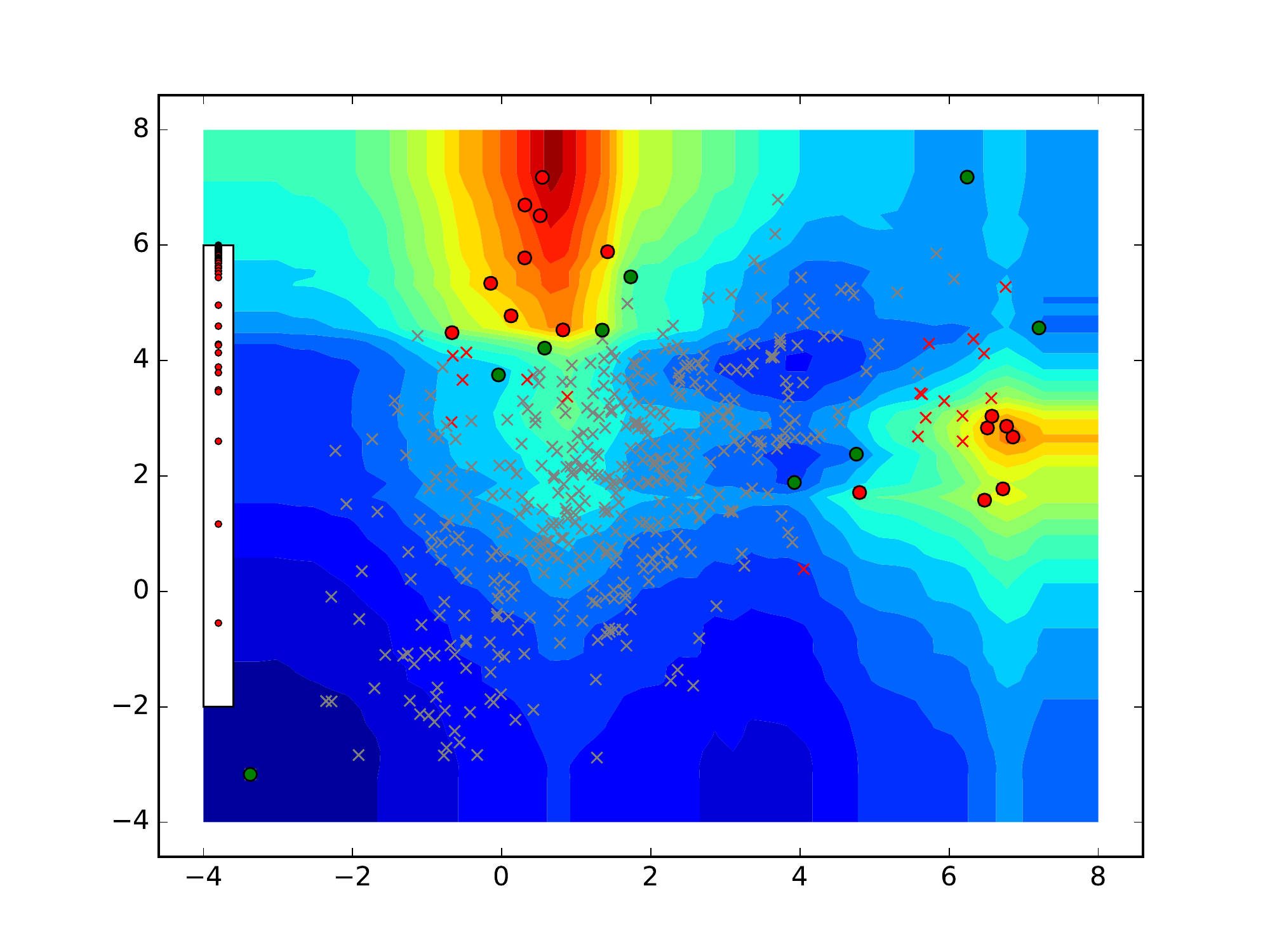}%
		\label{fig:ifor_iter_24}}
	\subfloat[32 Iterations]{\includegraphics[width=2.0in]{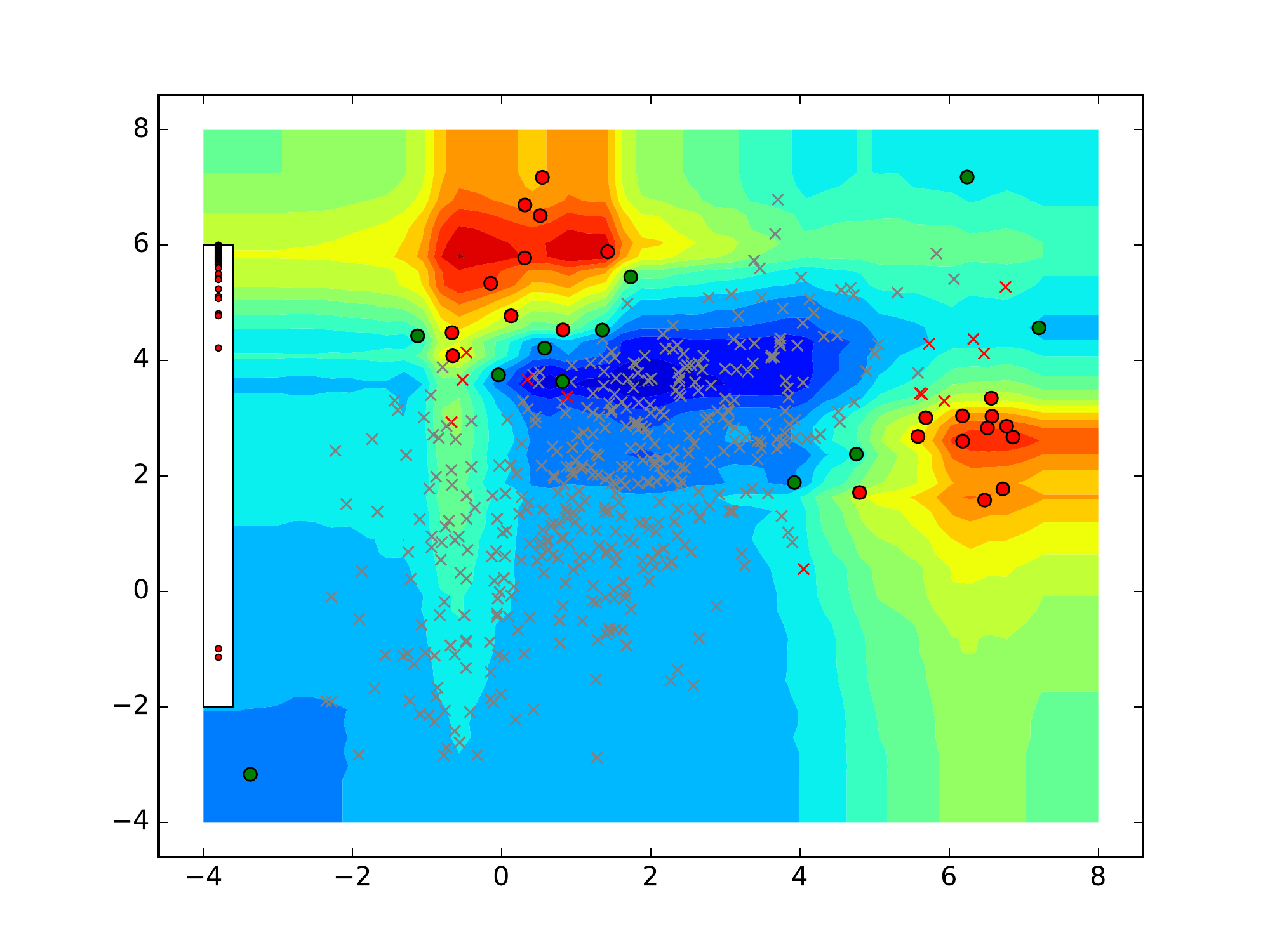}%
		\label{fig:ifor_iter_32}}
	\subfloat[Anomalies discovered]{\includegraphics[width=1.8in]{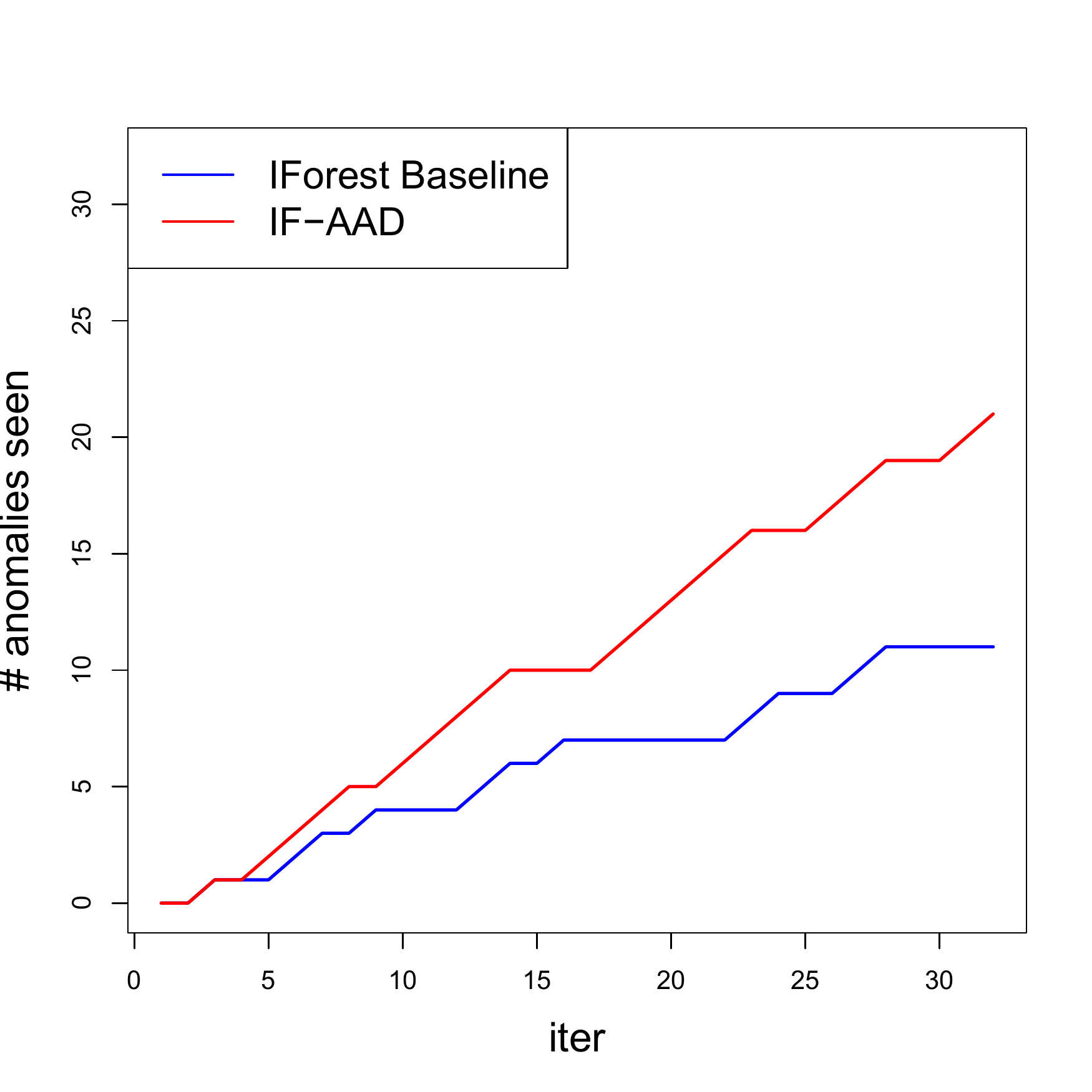}%
		\label{fig:ifor_synth_num_seen}}
	\caption{Incorporating feedback in Isolation Forest (IF) for synthetic data (Figure~\ref{fig:synthetic_data}). Figures~\ref{fig:ifor_iter_00} -- \ref{fig:ifor_iter_32} show anomaly score contours in the same way as explained in Figure~\ref{fig:ifor_contours_trees}. The \textcolor{red}{red} and \textcolor{britishracinggreen}{green} circles are the instances that have been presented for labeling. The x-axis in Figure~\ref{fig:ifor_synth_num_seen} represents the number of instances presented to the analyst, and the y-axis represents the number of true anomalies discovered. The \textcolor{red}{red} curve in Figure~\ref{fig:ifor_synth_num_seen} shows the number of true anomalies discovered when we incorporate feedback; the \textcolor{blue}{blue} curve in Figure~\ref{fig:ifor_synth_num_seen} shows the number of true anomalies discovered when no feedback was incorporated.}
	\label{fig:ifor_synth_feedback}
\end{figure*}

Our experiments consider starting with IF and tuning the weights based on feedback. This can simply be done by initializing the weights to all be constant values. The AAD algorithm can then be employeed with a regularization term that encourages weights to not depart too far from those initial values. 
We will refer to this algorithm as \textit{IF-AAD}. We assume that the forest is constructed exactly as in the original IF algorithm and the trees are kept fixed throughout the entire interaction with the analyst. 
That is, the feedback is employed only to re-weight the tree-partitions; the partitions themselves are never modified.

Figure~\ref{fig:ifor_synth_feedback} shows the result of incorporating feedback on the synthetic data. As the algorithm receives feedback, it alters the contours of the anomaly scores and focuses on the more relevant regions of the feature space. In all experiments we have set the number of trees $t = 100$. 
For the AAD parameters, we set $\tau=0.03$, and $C_A=100$, as recommended in \citet{Das:16}. We set $C_{\xi}=0.001$ in all experiments. A very large $C_{\xi}$ makes the algorithm focus more on regions where anomalies have already been found previously, and discourages exploration.

\section{Experiments}
\label{sec:experiments}
\begin{table*}[!t]
	\caption{Datasets used in our experiments, along with their characteristics.\label{tab:datasets}}
	{\begin{tabular}{llllll}
			\multicolumn{1}{l}{\bf Dataset} &\multicolumn{1}{l}{\bf Nominal Class} &\multicolumn{1}{l}{\bf Anomaly Class} &\multicolumn{1}{l}{\bf Total } & \multicolumn{1}{l}{\bf Dims } &\multicolumn{1}{l}{\bf \# Anomalies(\%)}
			\\ \hline \\
			Abalone         & 8, 9, 10 & 3, 21 & 1920 & 9 & 29 (1.5\%) \\
			\hline \\
			ANN-Thyroid-1v3         & 3 & 1 & 3251 & 21 & 73 (2.25\%) \\
			\hline \\
			Cardiotocography             & 1 (Normal) & 3 (Pathological) & 1700 & 22 & 45 (2.65\%) \\
			\hline \\
			Covtype             & 2 & 4 & 286048 & 54 & 2747 (0.9\%) \\
			\hline \\
			KDD-Cup-99             & \textit{`normal'} & \textit{`u2r', `probe'} & 63009 & 91 & 2416 (3.83\%) \\
			\hline \\
			Mammography             & -1 & +1 & 11183 & 6 & 260 (2.32\%) \\
			\hline \\
			Shuttle             & 1 & 2, 3, 5, 6, 7 & 12345 & 9 & 867 (7.02\%) \\
			\hline \\
			Yeast             & \textit{CYT}, \textit{NUC}, \textit{MIT} & \textit{ERL}, \textit{POX}, \textit{VAC} & 1191 & 8 & 55 (4.6\%) 
			\\ \hline
	\end{tabular}}
\end{table*}

In our experiments, we used the \textit{Mammography} \cite{woods:1993} dataset as well as seven datasets from the UCI repository \cite{uci}: \textit{Abalone}, \textit{Cardiotocography}, \textit{Thyroid (ANN-Thyroid)}, \textit{Forest Cover (Covtype)}, \textit{KDD-Cup-99}, \textit{Shuttle} and \textit{Yeast}. For each dataset, the classes were divided into two sets, one representing the nominal instances and a smaller set representing the anomlous instances. For the \textit{Cardiotocography} dataset, we retained all instances from the \textit{nominal} class as in the original dataset, but down-sampled the \textit{anomaly} instances so that they represent only around 2\% of the total data. The rest of the datasets were used in their entirety. The number of true anomalies and true nominals in each dataset along with the division of classes into nominals and anomalies are shown in Table~\ref{tab:datasets}. 

We evaluate an anomaly detector based on the rate that a simulated analyst is able to find true anomalies. In particular, each iteration of anomaly detection involves giving the analyst the top ranked instance and then receiving the feedback as \emph{anomalous} or \emph{nominal}. We compare our  proposed algorithm, IF-AAD, against the following baselines:
\begin{enumerate}
\item\textbf{IForest Baseline:} For the baseline, we present instances in decreasing order of anomaly score computed with the IF algorithm with uniform weights. This algorithm ignores the analyst feedback and thus the ranking is constant across iteration. This baseline captures the performance of an unsupervised anomaly detector that does not incorporate expert feedback. The trees were constructed by the original IF implementation available as part of the \textit{Python} \textit{scikit-learn} library.
\item\textbf{LODA-AAD:} This corresponds to the original AAD approach \cite{Das:16}, where AAD was applied to the ensemble of anomaly detectors created by the LODA anomaly detector \cite{pevny:2015}. Each anomaly detector in the ensemble corresponds to a random projection that maps each instance to 1D, bins the data to form a histogram, and then measures the anomaly score according to frequency of the histogram bin an instance falls into. 
\end{enumerate}

\begin{figure*}[!t]
	\centering
	\subfloat[Abalone]{\includegraphics[width=2.0in]{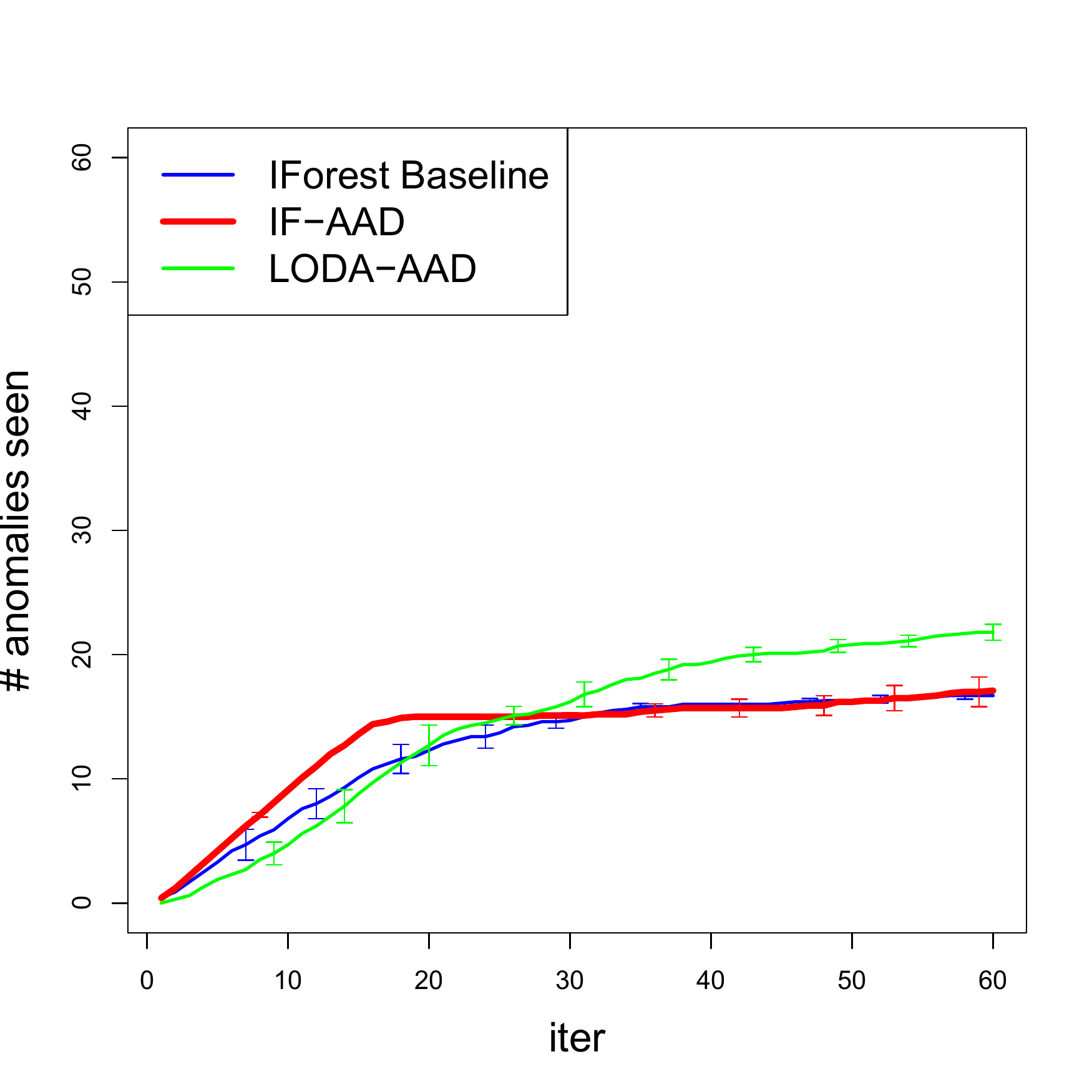}%
		\label{fig:iforest_abalone}}
	\subfloat[ANN-Thyroid-1v3]{\includegraphics[width=2.0in]{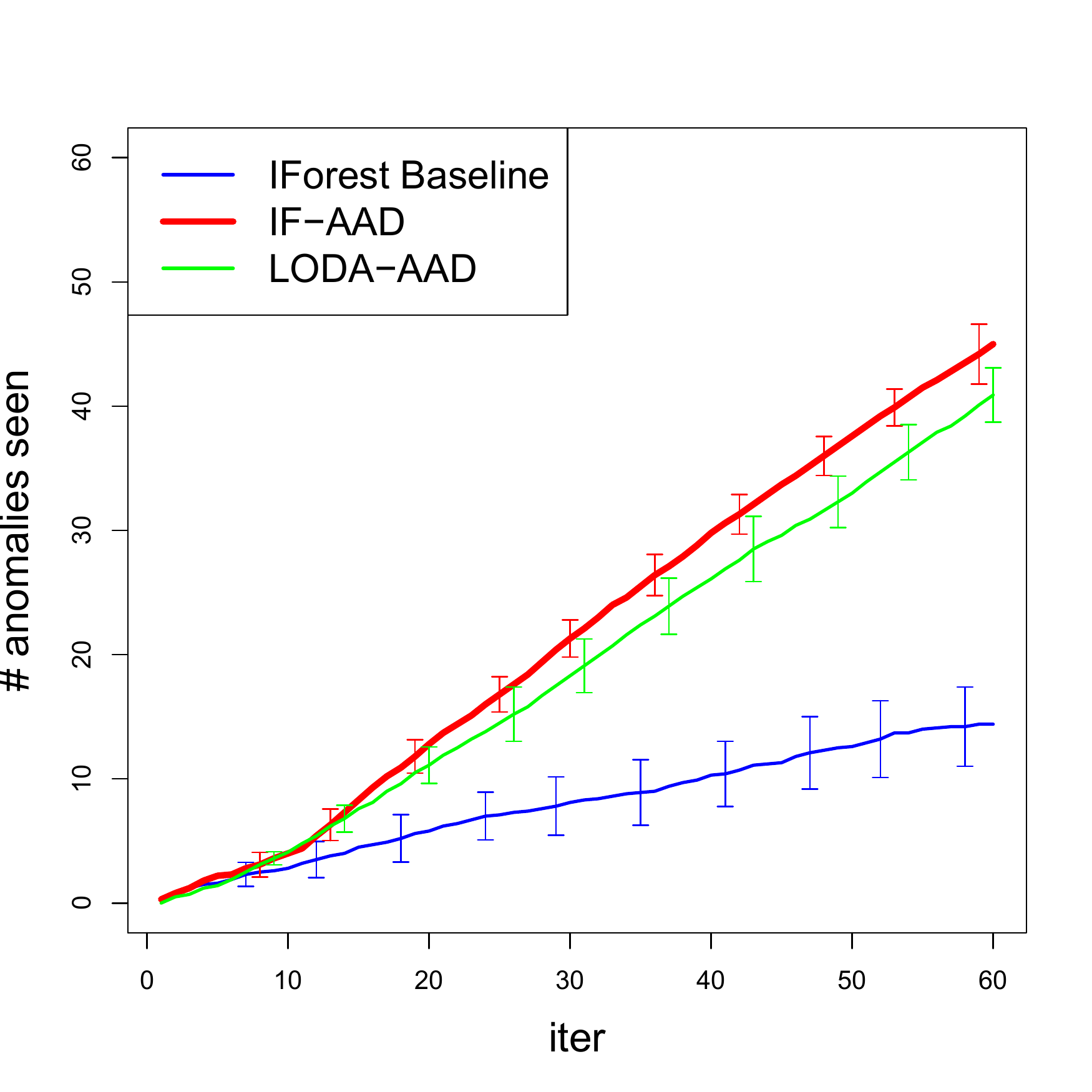}%
		\label{fig:iforest_ann_1v3}}
	\subfloat[Cardiotocography]{\includegraphics[width=2.0in]{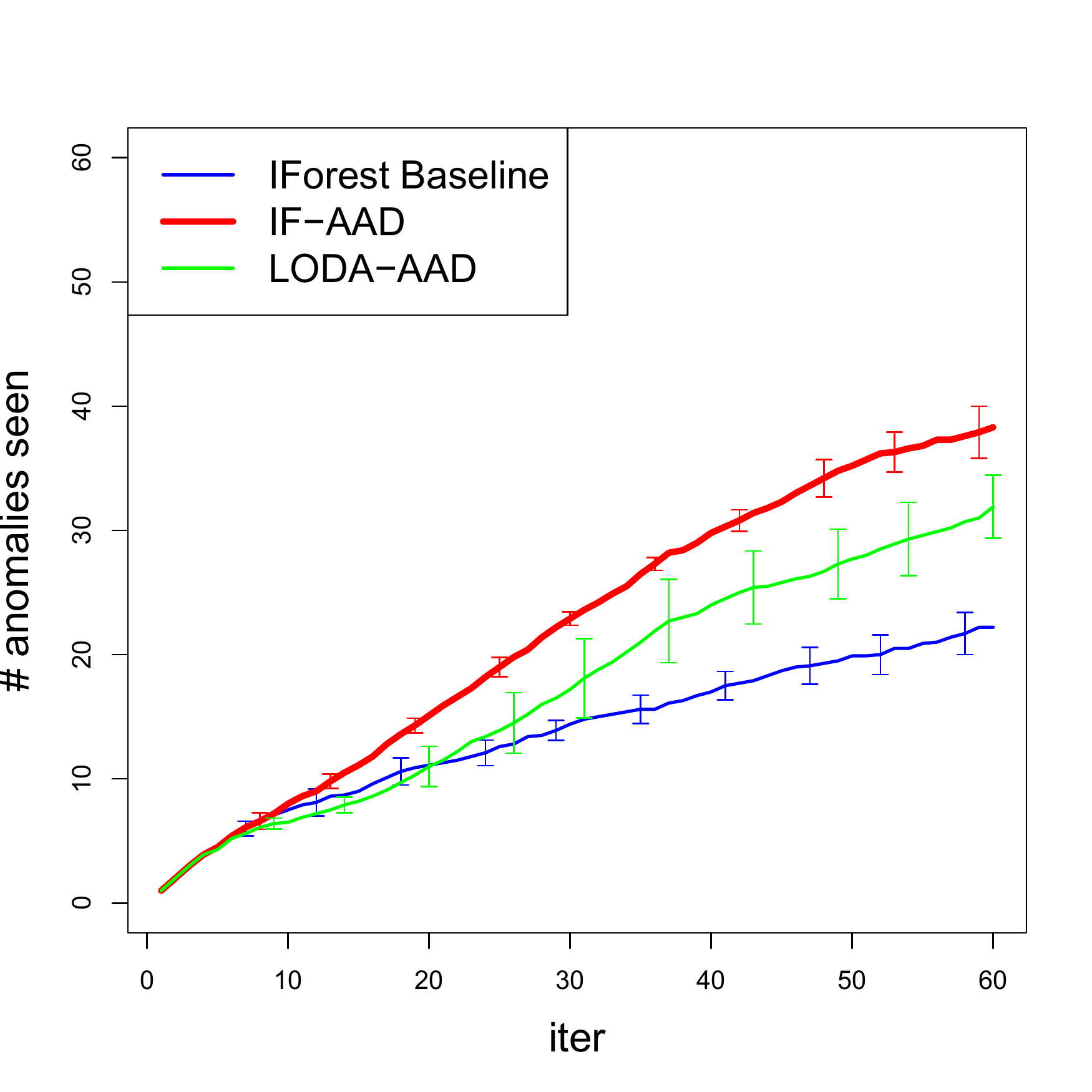}%
		\label{fig:iforest_cardiotocography}} \\[-3ex]
	\subfloat[Yeast]{\includegraphics[width=2.0in]{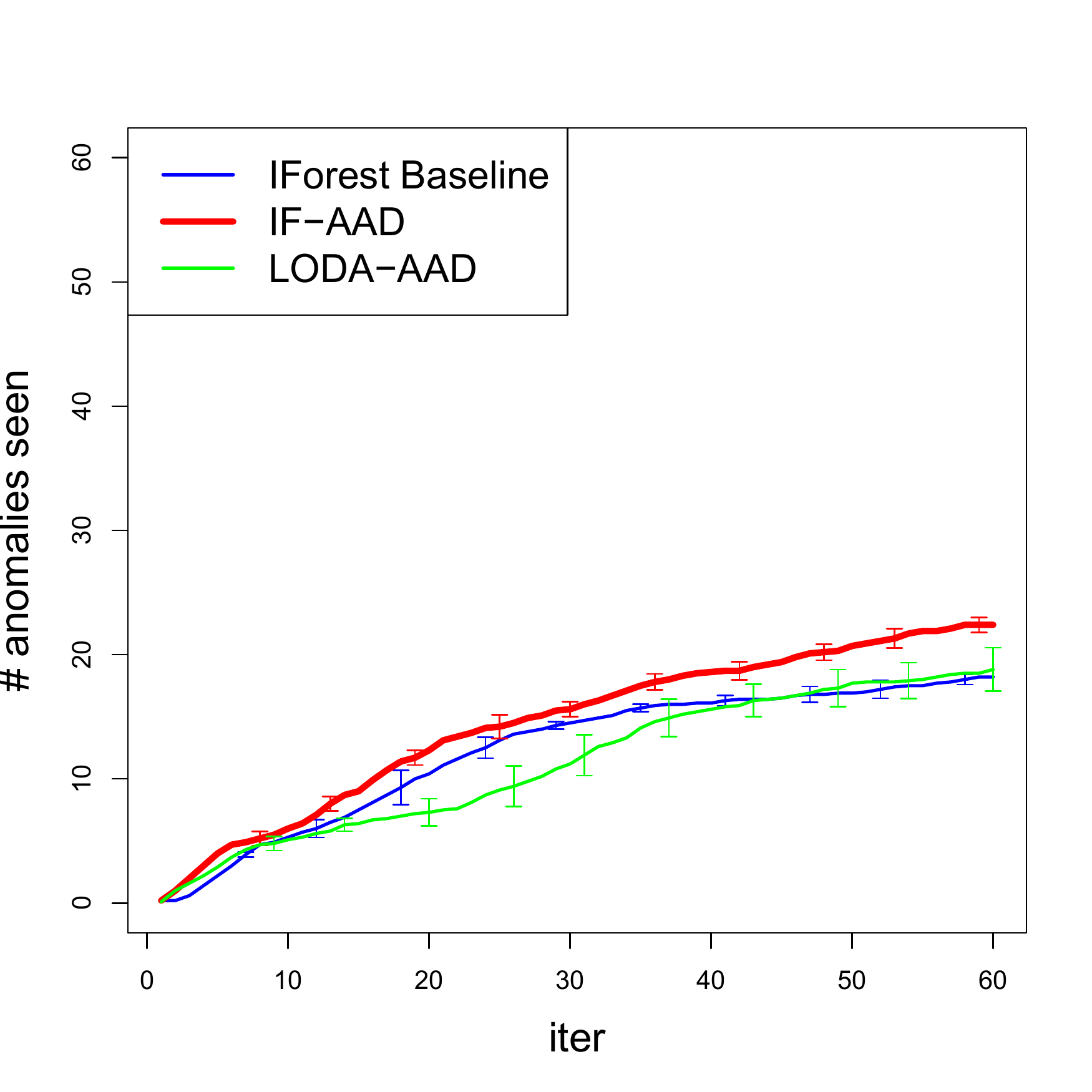}%
		\label{fig:iforest_yeast}}
	\subfloat[Covtype]{\includegraphics[width=2.0in]{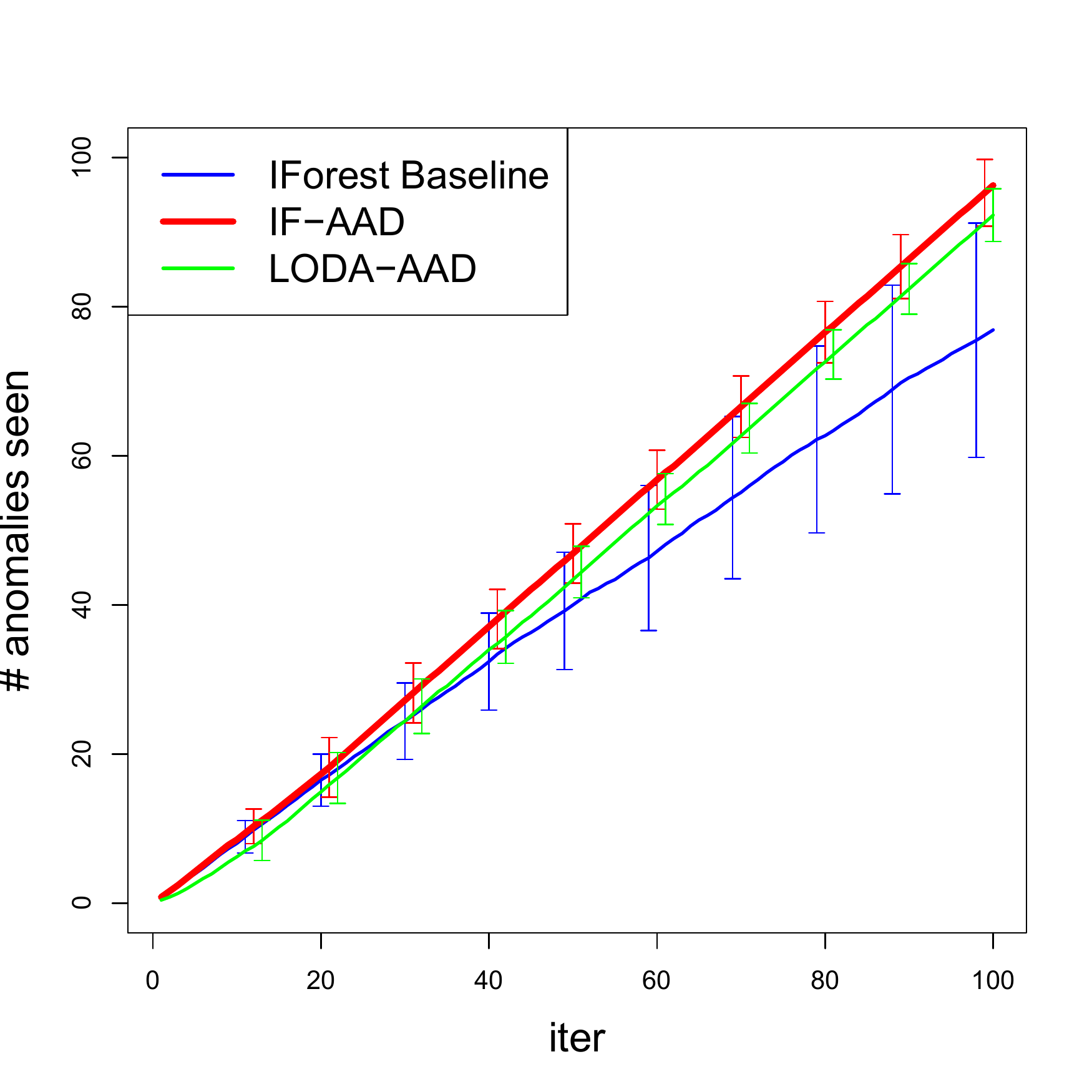}%
		\label{fig:iforest_covtype}}
	\subfloat[KDD-Cup-99]{\includegraphics[width=2.0in]{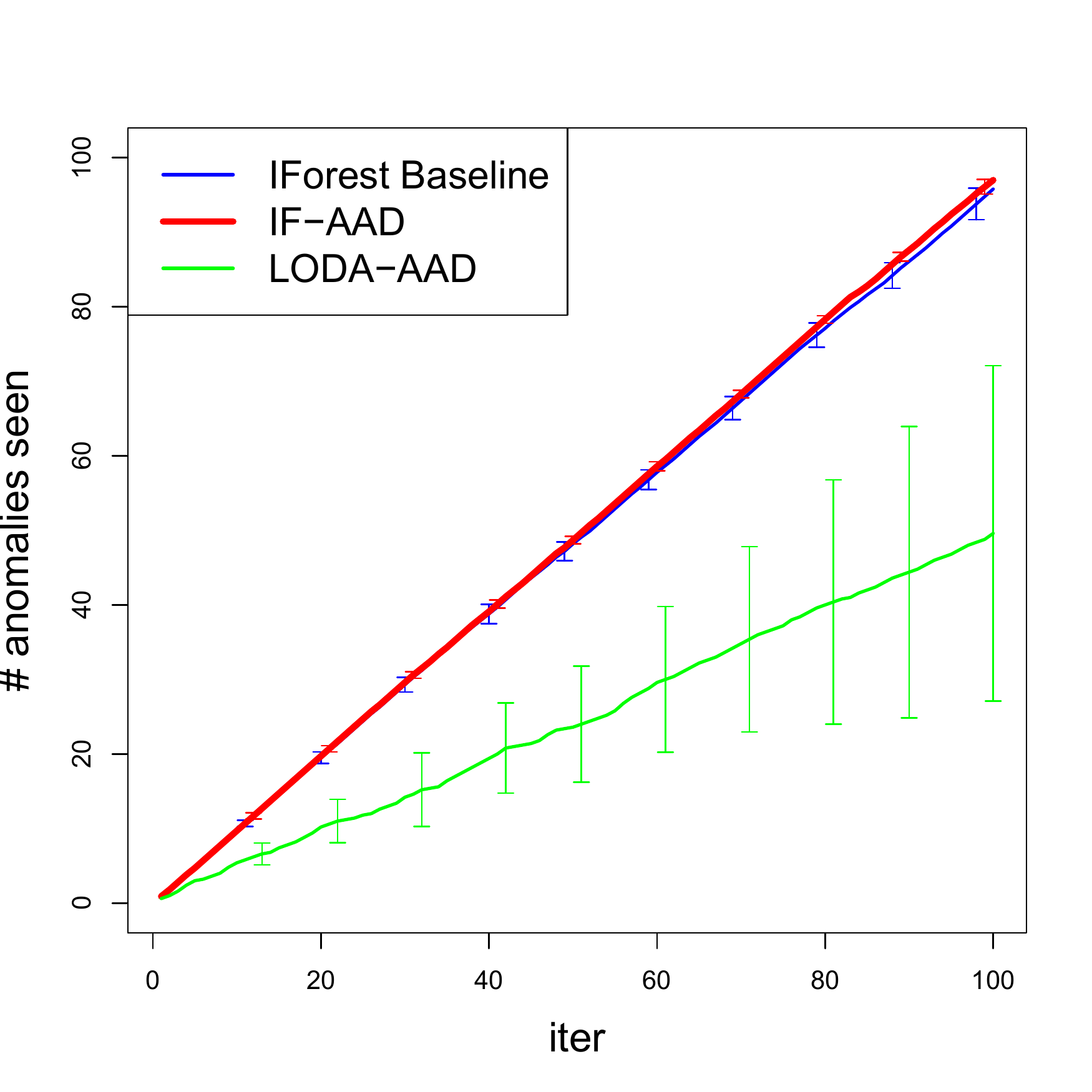}%
		\label{fig:iforest_kddcup}} \\[-3ex]
	\subfloat[Mammography]{\includegraphics[width=2.0in]{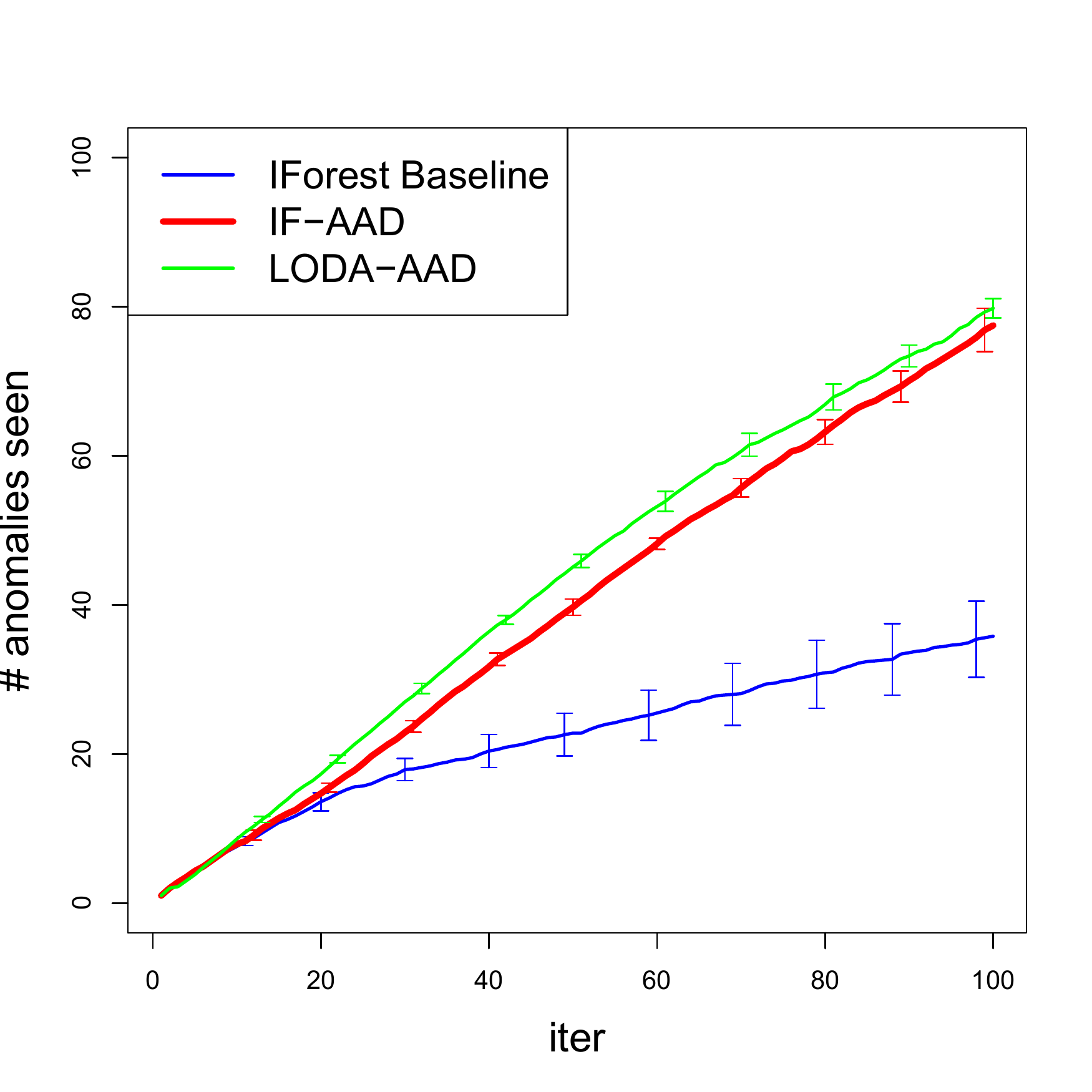}%
		\label{fig:iforest_mammography}}
	\subfloat[Shuttle]{\includegraphics[width=2.0in]{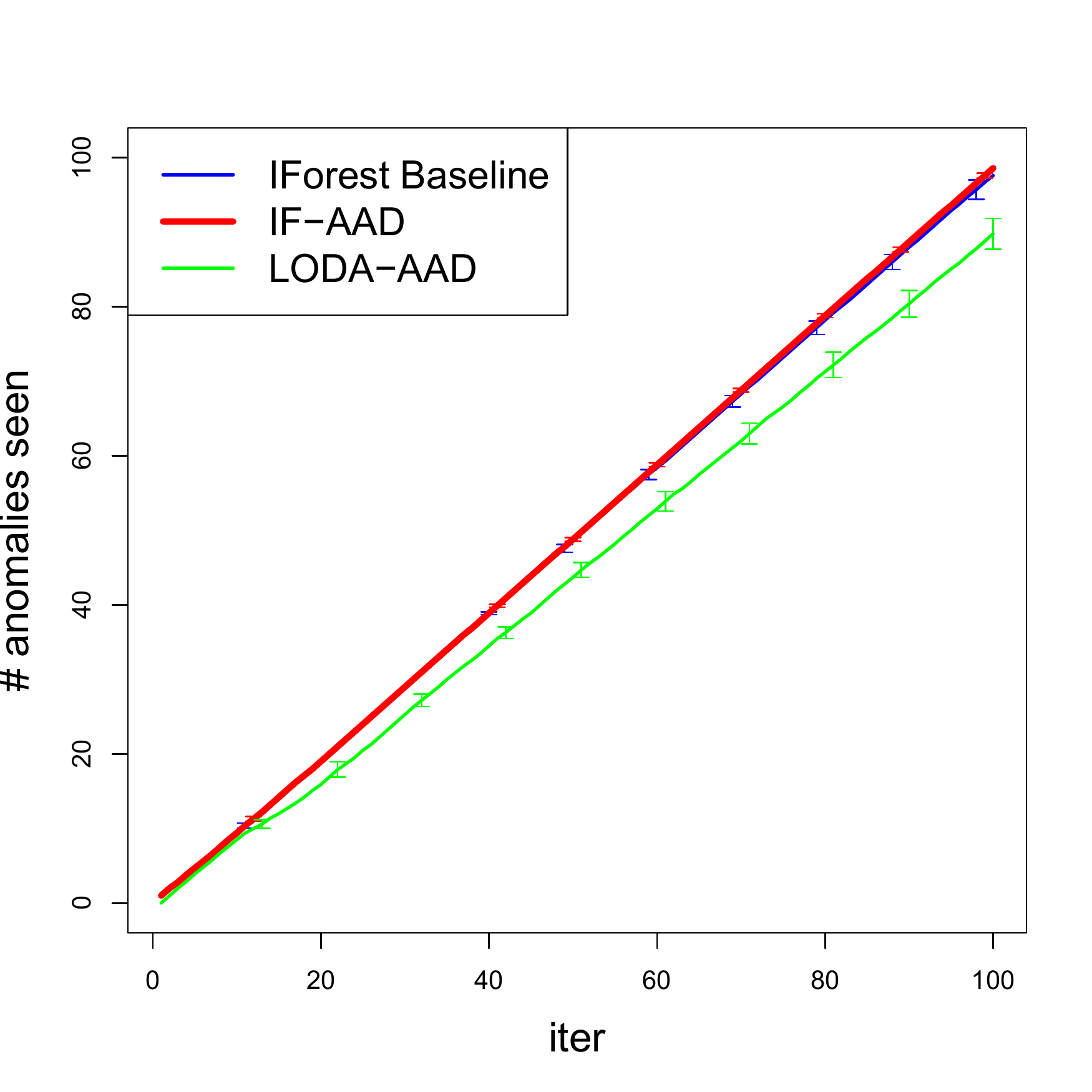}%
		\label{fig:iforest_shuttle}}
	\caption{The total number of true anomalies seen vs. the number of queries for all datasets. Total number of queries for the \textbf{smaller} datasets (\textit{Abalone}, \textit{Cardiotocography}, \textit{ANN-Thyroid-1v3}, and \textit{Yeast}) is $60$. Total number of queries for the \textbf{larger} datasets (\textit{Covtype}, \textit{KDD-Cup-99}, \textit{Mammography}, and \textit{Shuttle}) is $100$. Results were averaged over 10 runs. The error-bars represent $95\%$ confidence intervals.}
	\label{fig:iforest_all}
\end{figure*}

Figure~\ref{fig:iforest_all} shows the quantitative results for all of the data sets. Each graph plots the number of discovered anomalies versus the number of iterations. The best possible result is a line with slope 1, indicating that an anomaly is discovered at each iteration. The curves are averaged over 10 independent runs of the algorithm and 95\% confidence intervals are shown. Overall, we see that IF-AAD never hurts the performance of IF and in most cases significantly increases the number of anomalies discovered over time compared to both IF and LODA-AAD. 

\begin{figure*}[!t]
	\centering
	\subfloat[Abalone baseline]{\includegraphics[width=2.0in]{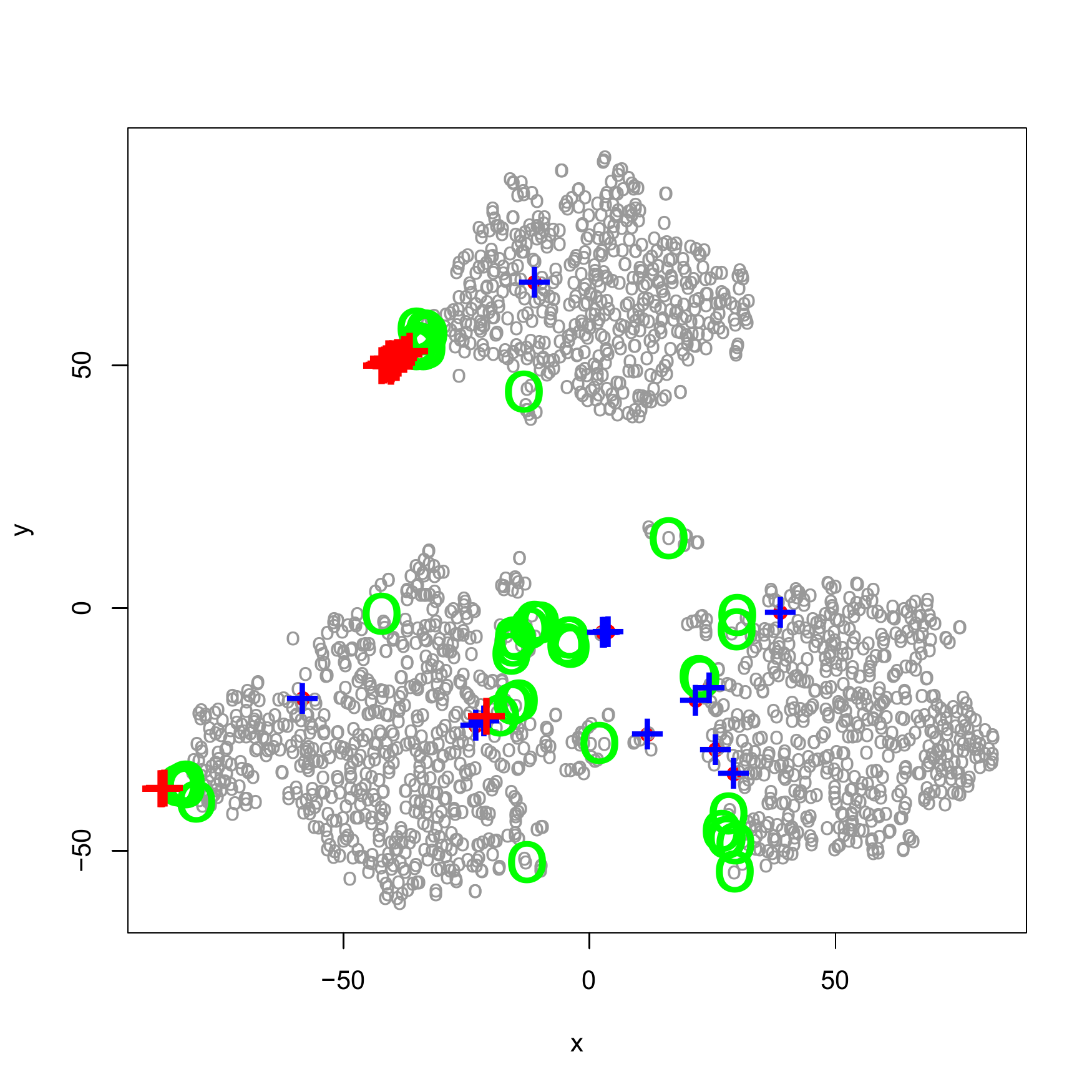}%
		\label{fig:tsne_iforest_abalone_baseline}}
	\subfloat[ANN-Thyroid-1v3 baseline]{\includegraphics[width=2.0in]{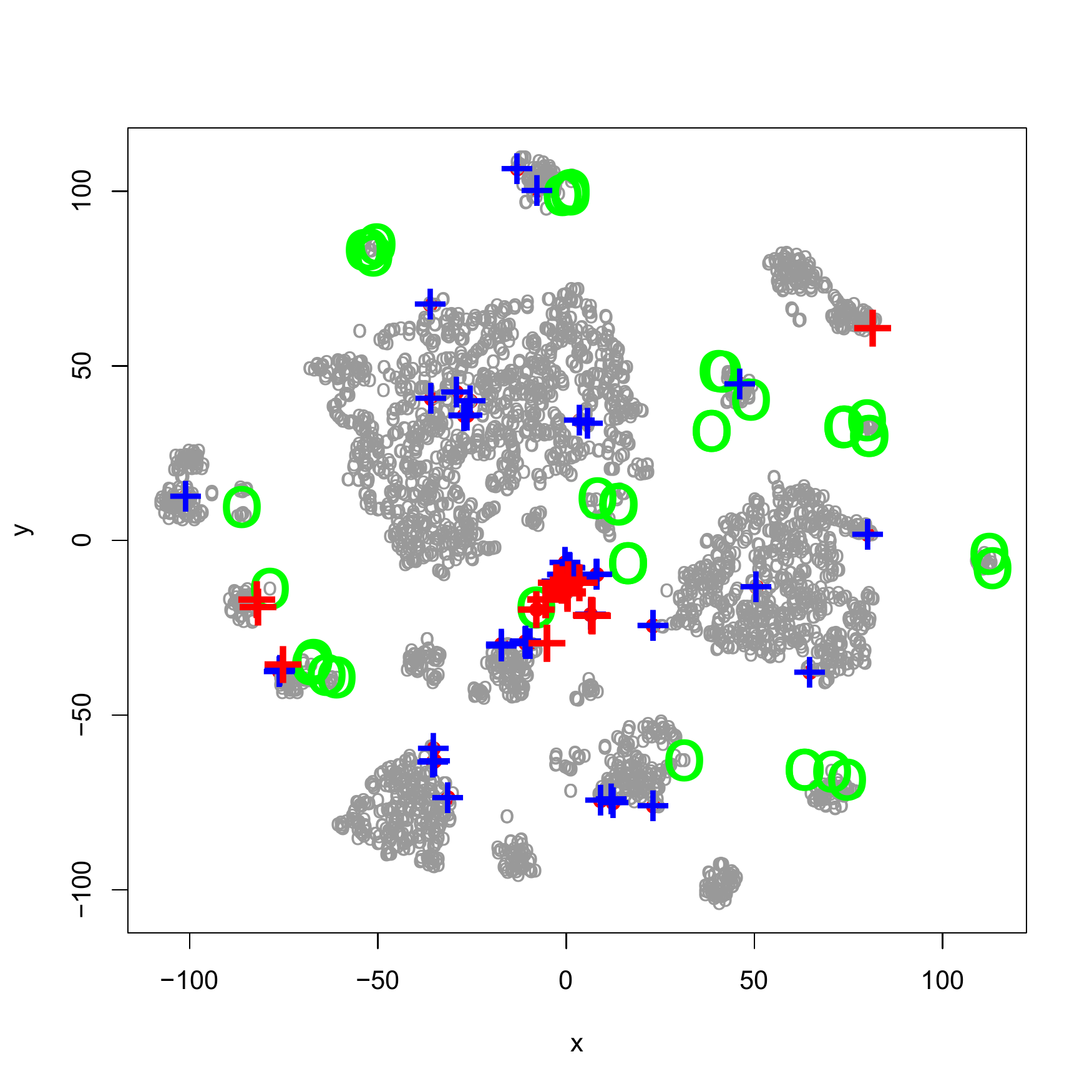}%
		\label{fig:tsne_iforest_ann_1v3_baseline}} \\
	\subfloat[Abalone IF-AAD]{\includegraphics[width=2.0in]{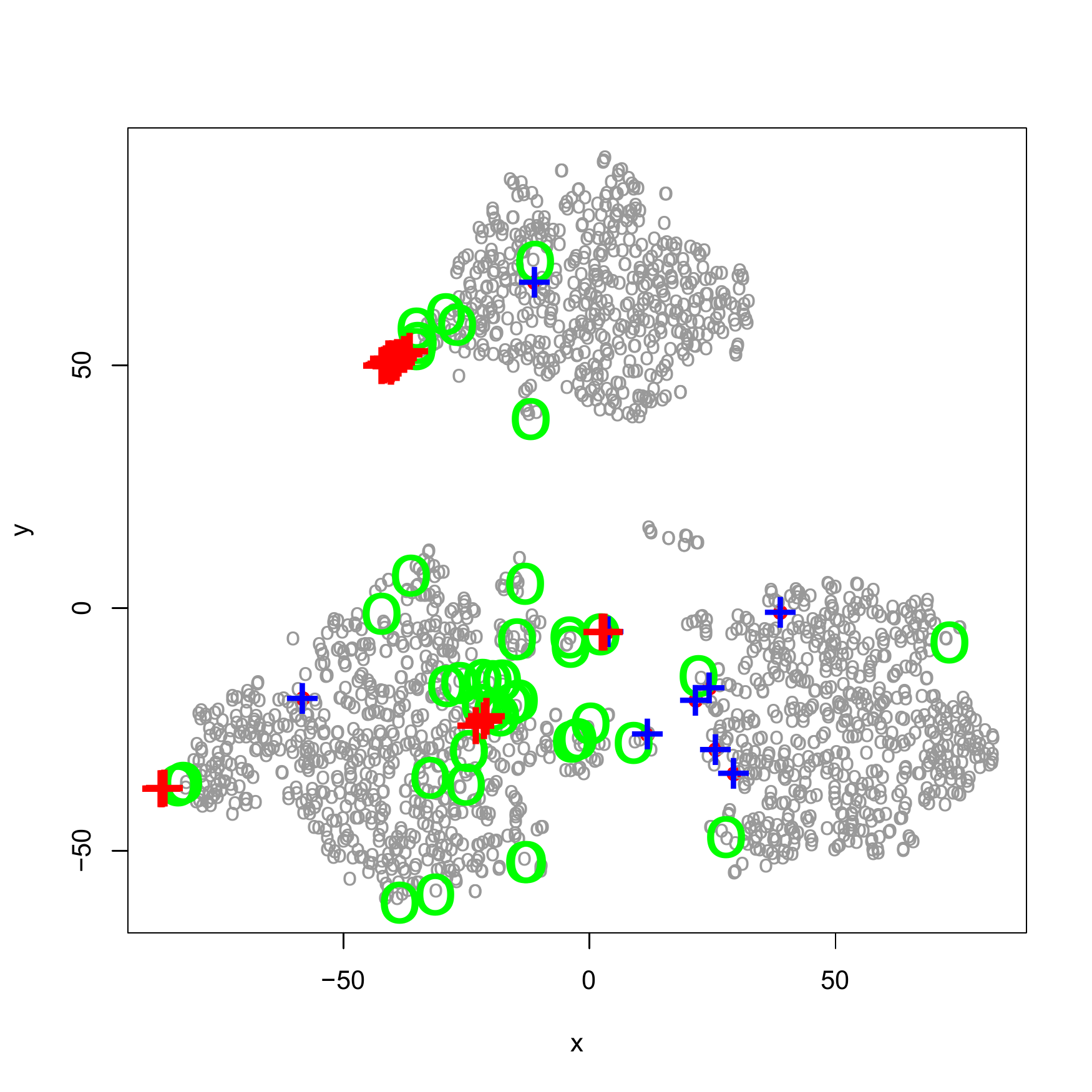}%
		\label{fig:tsne_iforest_abalone}}
	\subfloat[ANN-Thyroid-1v3 IF-AAD]{\includegraphics[width=2.0in]{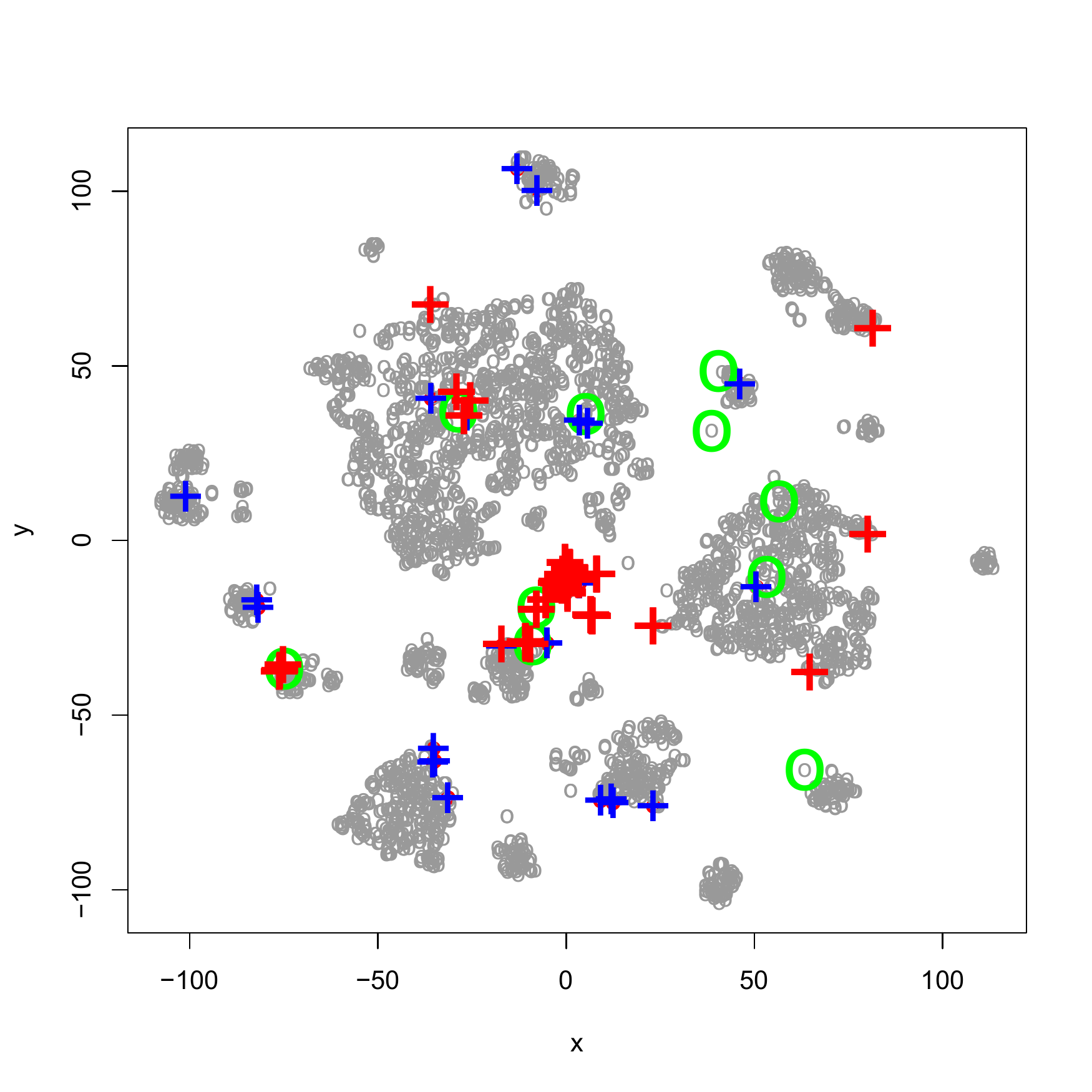}%
		\label{fig:tsne_iforest_ann_1v3}}
	\caption{Low-dimensional visualization of \textit{Abalone} and \textit{ANN-Thyroid-1v3} using t-SNE. Plus signs are anomalies and circles are nominals. A \textcolor{red}{red} coloring indicates that a true anomaly point was queried. A \textcolor{green}{green} indicates a nominal point was queried. Grey circles correspond to unqueried nominals. To make unqueried anomalies stand out visually, we indicate them with \textcolor{blue}{blue} plus signs.}
	\label{fig:iforest_tsne}
\end{figure*}

In order to gain more insight into how the feedback influences the algorithm on real-world datasets, we computed the two-dimensional representations of the datasets with \textit{t-SNE} \cite{maaten:2008} for visualization. Figure~\ref{fig:iforest_tsne} shows the t-SNE plots of two representative datasets, \textit{Abalone} and \textit{ANN-Thyroid-1v3}. We then marked the points on which the algorithm focused its queries in the first $60$ feedback iterations. We observe two ways by which the feedback influenced the queries. First, it reduced focus on the regions where the queried outliers were labeled nominal (e.g., location $(30, -50)$ in \textit{Abalone}, and $(60, -60)$ in \textit{ANN-Thyroid-1v3}). Second, it increased focus on regions that contained previously labeled true anomalies (e.g., $(-20, -20)$ in \textit{Abalone} and $(0, -10)$ in \textit{ANN-Thyroid-1v3}).

\begin{figure*}[!t]
	\centering
	\subfloat[ANN-Thyroid-1v3]{\includegraphics[width=2.0in]{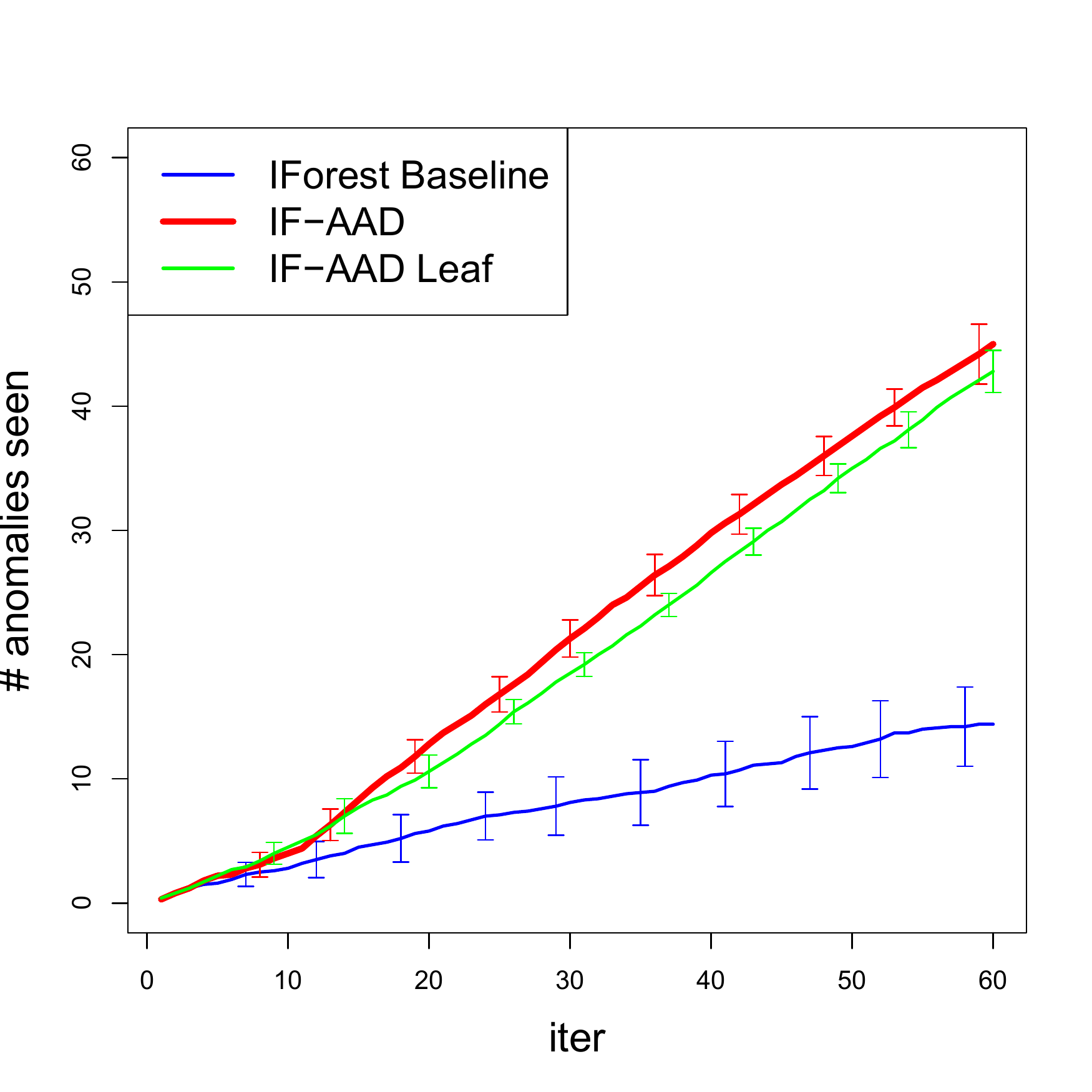}%
		\label{fig:iforest_leaf_ann_1v3}}
	\subfloat[Cardiotocography]{\includegraphics[width=2.0in]{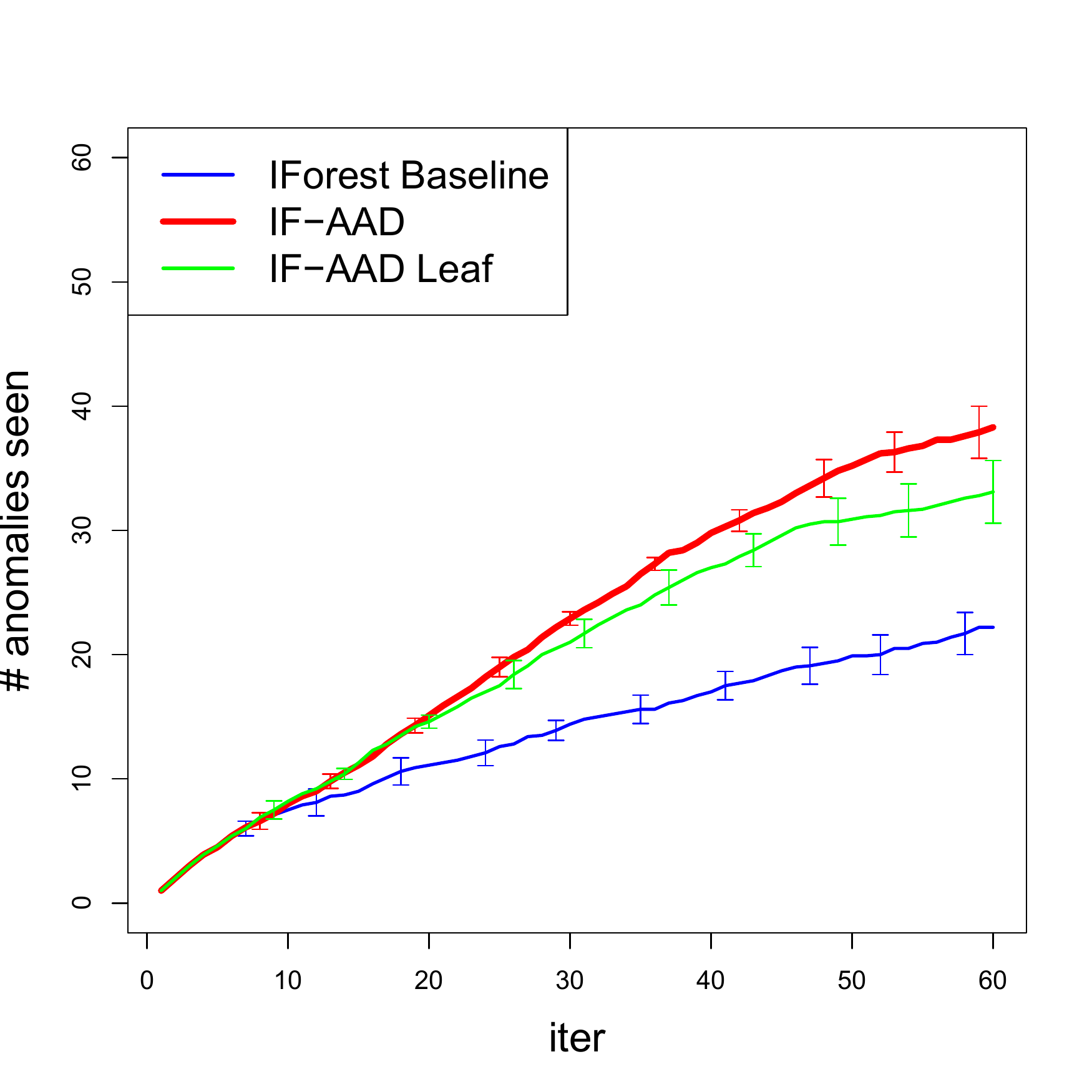}%
		\label{fig:iforest_leaf_cardiotocography}}
	\subfloat[Mammography]{\includegraphics[width=2.0in]{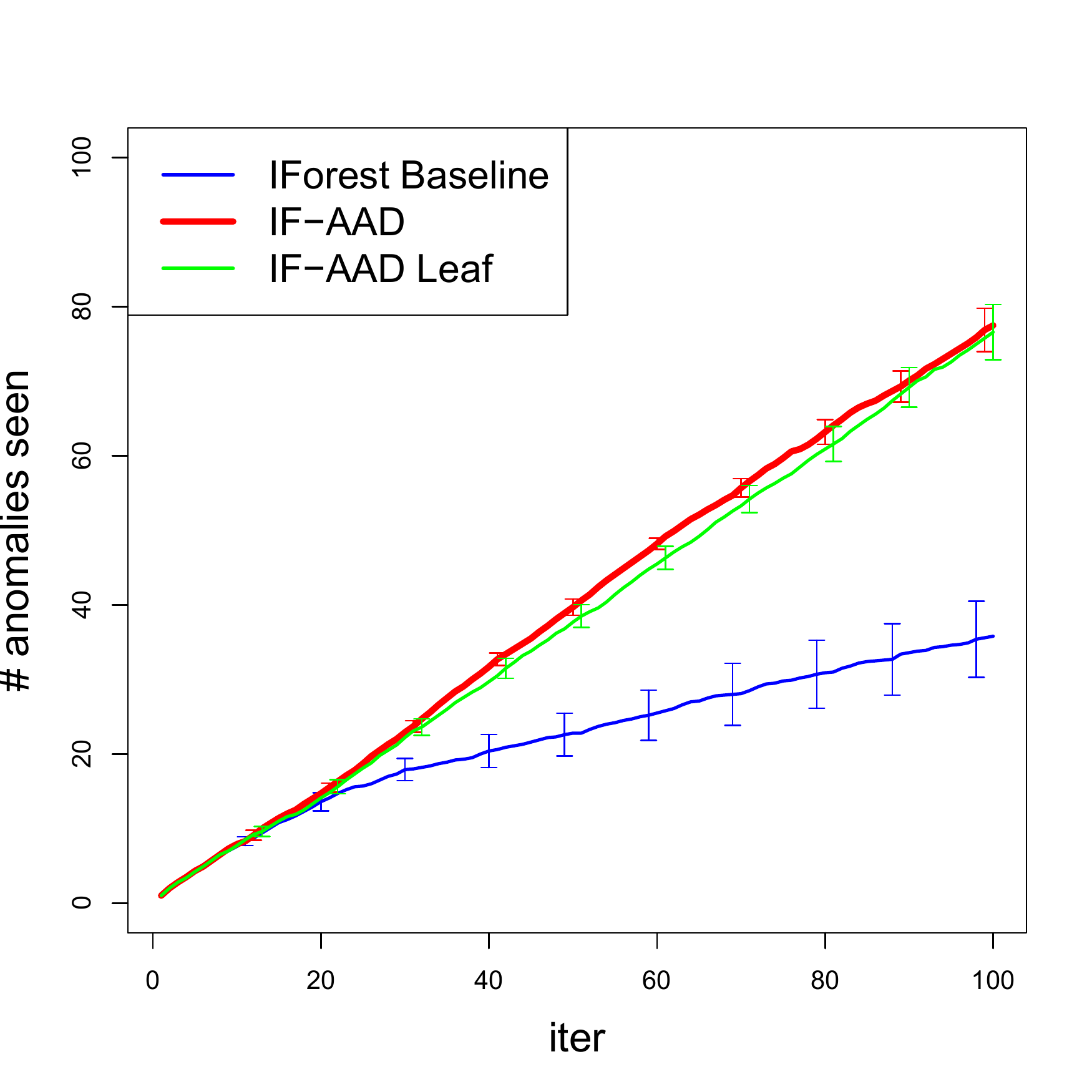}%
		\label{fig:iforest_leaf_mammography}}
	\caption{Comparison between assigning weights only at the leaf nodes (IF-AAD Leaf), and assigning weights at both the leaf and the intermediate nodes (IF-AAD). The curves show the total number of true anomalies seen vs. the number of queries. The weight at each leaf node in \textit{IF-AAD Leaf} was set to be the negative of the path length from the root, while the intermediate nodes were ignored. The weight at each node in \textit{IF-AAD} (leaf and intermediate) was set to $-1$.}
	\label{fig:iforest_leaf}
\end{figure*}

The time taken by IF-AAD in each feedback iteration depends on the particular data set and increases linearly with the number of labeled instances. As an example, for \textit{ANN-Thyroid-1v3}, IF-AAD took less than one second for the first feedback which involved one labeled instance, and took approx. 40 seconds to incorporate 100 labeled instances.

Finally, we note that a number of tree-based anomaly detectors are based on having non-zero weights only at the leaves (see Table 1). In order to evaluate the importance of having non-zero weights on internal nodes, we evaluated a version of IF-AAD that keeps all weights equal to zero except for the leaf nodes, which are updated by AAD. This new algorithm is called IF-AAD-Leaf and is implemented by only including indicator features and weights for leaf nodes in our formulation.  Figure~\ref{fig:iforest_leaf} shows a comparison between IF-AAD and IF-AAD-Leaf on three data sets that are representative of the results across all data sets. We observed that IF-AAD-Leaf has slightly worse performance than IF-AAD, showing that there is utility in weighting internal nodes, but the majority of the impact of feedback can be achieved by focusing just on leaf nodes.

\section{Summary}
\label{sec:conclusion}
We presented a new anomaly detection algorithm, IF-AAD, which fine-tunes the output of an Isolation Forest in a feedback loop. It treats the regions defined by the nodes of the isolation trees as components of an ensemble and re-weights them on the basis of feedback received from an analyst. IF-AAD is consistently one of the top performers in our experiments with real-world data. It sometimes detects twice the number of true anomalies as the baseline isolation forest algorithm. In future work we intend to extend our approach to other tree-based anomaly detectors.

\begin{acks}
	Funding was provided by Defense Advanced Research Projects Agency Contracts W911NF-11-C-0088 and FA8650-15-C-7557. The content of the information in this document does not necessarily reflect the position or the policy of the Government, and no official endorsement should be inferred. The U.S. Government is authorized to reproduce and distribute reprints for Government purposes notwithstanding any copyright notation here on. This paper is based upon work while Weng-Keen Wong was serving at the National Science Foundation. Any opinion, findings, and conclusions or recommendations expressed in this material are those of the authors and do not necessarily reflect the views of the National Science Foundation.
\end{acks}

\bibliographystyle{ACM-Reference-Format}
\bibliography{arpad} 

\end{document}